\documentclass{article}

\usepackage[square,sort,comma,numbers]{natbib}
\usepackage[preprint]{neurips_2020}

\usepackage[utf8]{inputenc} 
\usepackage[T1]{fontenc}    
\usepackage{hyperref}       
\usepackage{url}            
\usepackage{booktabs}       
\usepackage{amsfonts}       
\usepackage{nicefrac}       
\usepackage{microtype}      

\usepackage{times}
\usepackage{soul}
\usepackage{url}
\usepackage{color}
\usepackage[utf8]{inputenc}
\usepackage{caption}
\usepackage{graphicx}
\usepackage{amsmath}
\usepackage{amssymb}
\usepackage{amsthm}
\usepackage[noend]{algpseudocode}
\usepackage{booktabs}
\usepackage{algorithm}
\usepackage{subcaption}
\usepackage{wrapfig}

\newtheorem{thm}{Theorem}
\newtheorem{lem}{Lemma}

\newtheorem{cor}{Corollary}

\newtheorem{assumption}{Assumption}

\newcommand{\cN}{\mathcal{N}}

 \def\cB{{\mathcal{B}}} \def\cC{{\mathcal{C}}} \def\cD{{\mathcal{D}}}
\def\cE{{\mathcal{E}}}  \def\cG{{\mathcal{G}}} 
   
 \def\cN{{\mathcal{N}}} \def\cO{{\mathcal{O}}} \def\cP{{\mathcal{P}}}
  \def\cS{{\mathcal{S}}} 
 \def\cV{{\mathcal{V}}}

\def\ba{{\mathbf{a}}} \def\bb{{\mathbf{b}}}

  \def\bw{{\mathbf{w}}} \def\bx{{\mathbf{x}}} \def\by{{\mathbf{y}}}

\def\bA{{\mathbf{A}}}    
 \def\bG{{\mathbf{G}}}  \def\bI{{\mathbf{I}}} 
    
\def\bP{{\mathbf{P}}}    
  \def\bW{{\mathbf{W}}} \def\bX{{\mathbf{X}}}

     \def\d4{\!\!\!\!}

  \def\R{{\mathbb{R}}}

  \def\-{\! - \!}  \def\+{\! + \!}  \def\={\! = \!}  \def\>{\! > \!}

\newtheorem{remark}{Remark}

\newcommand{\bef}{\begin{figure}}
\newcommand{\eef}{\end{figure}}
\newcommand{\beq}{\begin{eqnarray}}
\newcommand{\eeq}{\end{eqnarray}}

\usepackage[utf8]{inputenc}

\def\arg{\mathrm{arg}}
\def\Var{\mathrm{Var}}

\title{Straggler-Resilient Distributed Machine Learning with Dynamic Backup Workers}

\author{
Guojun Xiong\thanks{Equal contribution.  Ordering determined by alphabetical order.}\\
SUNY-Binghamton University\\
Binghamton, NY 13902\\
 \texttt{gxiong1@binghamton.edu}
  \And
  Gang Yan$^*$\\
SUNY-Binghamton University\\
Binghamton, NY 13902\\
 \texttt{gyan2@binghamton.edu}
  \And
  Rahul Singh\\
Indian Institute of Science\\
Bengaluru, Karnataka 560012, India\\
 \texttt{rahulsingh@iisc.ac.in}
  \And
Jian Li\\
SUNY-Binghamton University\\ 
Binghamton, NY 13902\\
 \texttt{lij@binghamton.edu}
}

\begin{document}

\maketitle

\begin{abstract}

With the increasing demand for large-scale training of machine learning models, consensus-based distributed optimization methods have recently been advocated as alternatives to the popular parameter server framework.  In this paradigm, each worker maintains a local estimate of the optimal parameter vector, and iteratively updates it by waiting and averaging all estimates obtained from its neighbors, and then corrects it on the basis of its local dataset. However, the synchronization phase can be time consuming due to the need to wait for \textit{stragglers}, i.e., slower workers.  An efficient way to mitigate this effect is to let each worker wait only for updates from the fastest neighbors before updating its local parameter.  The remaining neighbors are called \textit{backup workers.}  To minimize the globally training time over the network, we propose a fully distributed algorithm to dynamically determine the number of backup workers for each worker.  We show that our algorithm achieves a linear speedup for convergence (i.e., convergence performance increases linearly with respect to the number of workers). We conduct extensive experiments on MNIST and CIFAR-10 to verify our theoretical results. 

\end{abstract}

\section{Introduction}\label{intro}

Highly over-parametrized deep neural networks have shown impressive results in a variety of machine learning (ML) tasks such as computer vision \cite{he2016deep}, natural language processing \cite{vaswani2017attention}, speech recognization \cite{amodei2016deep} and many others.  Its success depends on the availability of large amount of training data, which often leads to a dramatic increase in the size, complexity, and computational power of the training systems.  In response to these challenges, the need for efficient parallel and distributed algorithms (i.e., data-parallel mini-batch stochastic gradient descent (SGD))
becomes even more urgent for solving large scale optimization and ML problems \cite{bekkerman2011scaling,boyd2011distributed}. These ML algorithms in general
use the parameter server (PS) \cite{valiant1990bridging,smola2010architecture,dean2012large,ho2013more,li2014scaling,li2014communication} or {ring All-Reduce} \cite{gibiansky17allreduce,goyal2017accurate} communication primitive to perform exact averaging on the local mini-batch gradients computed on different subsets of the data by each worker, for the later synchronized model update.  However, the aggregation with PS or All-Reduce often leads to extremely high communication overhead\footnote{Each worker needs to receive the aggregate of updates from all other workers to move to the next iteration, where aggregation is performed either by PS or along the ring through multiple rounds.}, causing the bottleneck in efficient ML model training.  

Recently, an alternative approach has been developed in ML community \cite{lian2017can,lian2018asynchronous}, where each worker keeps updating a local version of the parameter vector and broadcasts its updates \textit{only} to its neighbors.  The popularity of this family of algorithms can date back to the seminal work of \cite{tsitsiklis1986distributed} on distributed gradient methods, in which, they are often referred to as \textit{consensus-based distributed optimization methods.}

\noindent\textbf{Dynamic Backup Workers.}  Most existing consensus-based algorithms assume full worker participation, that is, all workers participate in every training round.  Such an \textit{exact averaging} with consensus-based methods are sensitive to \textit{stragglers}, i.e., slow tasks, which can significantly reduce the computation speed in a multi-machine setting \cite{ananthanarayanan2013effective,karakus2017straggler}. In practice, only a small fraction of workers participate in each training round, thus rendering the active worker (neighbor) set stochastic and time-varying across training rounds.  An efficient way to mitigate the effect of stragglers is to rely on \textit{backup workers} \cite{dean13the}:  rather than waiting for updates from all neighbors (say $n_j$), worker $j$ only waits for updates from the fastest $p_j$ neighbors before updating and correcting its local parameter.  The remaining $b_j\triangleq n_j-p_j$ workers  are called backup workers.

Workers are then nodes of \textit{a time-varying consensus graph}, where an edge $(i,j)$ indicates that worker $j$ waits updates from worker $i$ to update and correct its local parameter.  A larger value of $b_j, \forall j$ reduces the communication overhead and may mitigate the effect of stragglers since each worker needs to wait for fewer updates from its neighbors.   This reduces the period of one iteration.  At the same time, the ``quality" of updates may suffer since each worker only receives limited information from neighbors.  As a result, more iterations will be required for convergence or in some cases, it cannot even guarantee the consistency of parameters across all workers.  This raises the questions: Can a large number of backup workers per node significantly reduce the convergence time or will stragglers still slow down the whole network? and which effect prevails?  Apart from some numerical results \cite{luo2019hop}, this paper is perhaps the first attempt to answer these questions.

\noindent\textbf{Main Contribution.}
Our contribution is three-fold.  First, we formulate the consensus-based distributed optimization problem with dynamic backup workers (DyBW), and propose the consensus-based DyBW (\texttt{cb-DyBW}) algorithm that can dynamically adapt the number of backup workers for each worker during the training time.  

Second, we present the first (to the best of our knowledge) convergence analysis of consensus-based algorithms with DyBW that is cognizant of training process at each worker.  We show that the convergence rate of \texttt{cb-DyBW} algorithm on both independent-identically-distributed (i.i.d.) and non-i.i.d. datasets across workers is $\mathcal{O}\left(\frac{1}{\sqrt{NK}}+\frac{1}{K}\right)$, where $N$ is the total number of workers and $K$ is the total number of communication rounds.  This indicates that our algorithm achieves a linear speedup for convergence rate for a sufficiently large $K.$  The PSGD algorithm \cite{lian2017can} also achieves the same rate with full worker participation (i.e., $b_j=0,\forall j$) in each communication round, which lead to high communication costs and implementation complexity.  In contrast we have the flexibility to dynamically define the number of backup workers for each node in the system. We then build on the result of convergence rate in terms of number of iterations to characterize \textit{the wall-clock time} required to achieve this precision.  We show that \texttt{cb-DyBW} can dramatically reduce the convergence time by reducing the duration of an iteration without increasing the number of convergence iterations for a certain accuracy.

Third, we develop a practical algorithm to dynamically determine the number of backup workers for each worker during the training time.
It appears that our paper is the first study to understand and quantify the effect of dynamic backup workers in fully distributed ML systems.

\noindent\textbf{Related Work.} Our algorithm belongs to the class of consensus/gossip algorithms.  A significant amount of research has focused on developing consensus-based algorithms in various fields including \cite{bertsekas1989parallel,kempe2003gossip,xiao2004fast,boyd2006randomized,nedic2009distributed,johansson2010randomized,ram2010distributed,boyd2011distributed,duchi2011dual,colin2016gossip,scaman2017optimal,scaman2018optimal,tang2018communication,tang2018d}.  However, most results are developed for full worker participation, which is communication expensive with large numbers of workers.  Efficient control of partial worker participation in consensus-based distributed optimization methods has received little attention, which arise in problems of practical interests. The few existing works address this issue through a stale-synchronous model \cite{chen2016revisiting,ho2013more} have no convergence guarantee, and the number of backup workers is often configured manually through some preliminary experiments before the start of the actual training process.  Adaptive worker selection that is cognizant of the training process at each worker has not been understood yet.

A more relevant line of research on distributed learning is the PS/All-Reduce.  Most existing work is based on the centralized framework through synchronous communication scheme which often leads to communication congestion in the server.  Alternatively, the PS can operate asynchronously, updating parameter vector immediately after it receives updates from a single worker.  Although such an asynchronous update manner can increase system throughput (parameter updates per time unit), some workers may still operate on stale versions of the parameter vector and in some cases, even preventing convergence to the optimal model \cite{dai2018toward}.
More recently, several PS-based algorithms considering stale-synchronous model via backup workers have gained increased attention.  See \cite{chen2016revisiting,teng2018bayesian,ruan2020towards,xu2020dynamic,yang2021achieving} for details, among which \cite{luo2019hop,xu2020dynamic} are perhaps the only other works proposing to dynamically adapt the number of backup workers.  However, most of them focus on showing empirical results without convergence guarantees and without a rigorous analysis of how selection skew affects convergence speed.  Moreover, none of them exploit the communication among workers, since they assume either a central server exists to coordinate workers (e.g., PS model) or all workers have an identical role (e.g., All-Reduce).  Generalizing the PS schemes to our fully distributed framework is a non-trivial and open problem because of the large scale distributed and heterogeneous nature of the training data.   We refer the interested readers to \cite{tang2020communication} and references therein for a comprehensive review in distributed learning.

\noindent\textbf{Notation.}
Let $N$ be the total number of workers and $K$ be the number of total communication rounds.  We denote the cardinality of a finite set $\mathcal{A}$ as $|\mathcal{A}|.$  We denote by $\bI_M$ and $\mathbf{1}$ ($\mathbf{0}$) the identity matrix and all-one (zero) matrices of proper dimensions, respectively.  We also use $[N]$ to denote the set of integers $\{1,\cdots, N\}$.  We use boldface to denote matrices and vectors, and $\|\cdot\|$ to denote the $l_2$-norm. In addition, $\text{diag}[\bA]$ returns the diagonal elements of matrix $\bA.$

\section{Background and Problem Formulation}
\label{prelim}

In this section, we introduce the background and formulation of the consensus-based distributed optimization problem with dynamic backup workers.

\subsection{Consensus-based Distributed Optimization}
Supervised learning aims to learn a function that maps an input to an output using $L$ examples from a training dataset $\cD=\{(\bx_\ell, y_\ell), \ell=1,\ldots, L\}$, where each example is a pair of input $\bx_\ell$ and the associated output $y_\ell$.  The training of ML model aims to find the best statistical model via optimizing a set of parameters $\bw\in\R^{d\times 1}$ to solve the following optimization problem 
\begin{equation}\label{eq:central_objective}
\min_\bw\quad \sum\limits_{\ell=1}^L f(\bw, \bx_\ell, y_\ell),
\end{equation}
where $f(\bw, \bx_\ell, y_\ell)$ is the model error on the $l$-th element of dataset $\cD$ when parameter $\bw$ is used.  The objective function may also include a regularization term that enforces some ``simplicity" (e.g., sparseness) of $\bx$, which can be easily taken into account in our analysis.

Different iterative algorithms have been proposed to solve~(\ref{eq:central_objective}) and we refer interested readers to \cite{bubeck2015convex} for a nice introduction.
Due to increases in available data set and complexity of statistical model, an efficient distributed algorithm for \eqref{eq:central_objective} is usually desired to determine the parameter vector in a reasonable time. The common way is to offload the computation overhead to $N$ independent workers, and they jointly determine the optimal parameter of interest through a distributed coordination. 
In other words, \eqref{eq:central_objective} can be equivalently reformulated as a minimization of the sum of functions local to each worker
\beq
\min_\bw\quad F(\bw)\triangleq \frac{1}{N}\sum\limits_{j=1}^N F_j(\bw),
 \eeq
where $F_j(\bw)\triangleq \frac{1}{|\cD_j|}\sum_{(\bx_\ell,y_\ell)\in\cD_j}f(\bw, \bx_\ell,y_\ell)$ and $\cD_j$ is the local dataset of worker $j\in[N]$.  In conventional distributed learning, $\cD$ is divided among $N$ workers ($\cD=\cup_{j=1}^{N}\cD_j$) and the distribution for each worker's local dataset can usually be assumed to be i.i.d.. Unfortunately, this assumption may not hold true in practice.  Instead, we make no so such assumption in our model and our analysis holds for both i.i.d. and non-i.i.d. local datasets across workers.

The distributed system can be modeled as \textit{a communication graph} $\cG=(\cN,\cE)$ with $\cN=[N]$ being the set of workers and an edge $(i, j)\in\cE$ indicates that workers $i$ and $j$ can communicate with each other.
Without loss of generality (W.l.o.g.), we assume the graph is strongly connected, i.e., there exists at least one path between any two arbitrary workers. Denote the neighbors of worker $j$ as $\cN_j=\{i|(i,j)\in\cE\}\cup\{j\}$. Worker $j$ maintains a local estimate of the parameter vector $\bw_j(k)$ at iteration $k$ and broadcasts it to its neighbors.
The local estimate is updated as follows:
\begin{equation}\label{eq:local-estimation-all}
\bw_j({k+1})=\sum_{i\in \cN_j} \bw_i(k)P_{i,j}-\eta_j(k) g(\bw_j(k)),
\end{equation}
where $\bP=(P_{i,j})$ is a $N\times N$ non-negative matrix and we call it the \textit{consensus matrix}.  Parameter $\eta(k)>0$ is the learning rate, which can be time-varying.  In other words, in each iteration, worker $j\in[N]$ computes a weighted average (i.e., consensus component) of the estimates of its neighbors and itself, and then corrects it by taking into account a stochastic subgradient $g(\bw_j(k))$ of its local function, i.e.,
\begin{equation}\label{eq:local-subgradient}
g(\bw_j(k))\triangleq\frac{1}{|\cC_j(k)|} \sum\limits_{(\bx_\ell, y_\ell)\in \cC_j(k)} \nabla f(\bw_j(k), \bx_\ell, y_\ell),
\end{equation}
where $\cC_j(k)$ is a random mini-batch of size $|\cC_j(k)|$ drawn from $\cD_j$ at iteration $k$.

\subsection{Dynamic Backup Workers Setting}

To mitigate the effect of stragglers, worker $j$ only waits for the first $p_j(k)$ updates from its neighbors rather than waiting for all as in~(\ref{eq:local-estimation-all}) at iteration $k$.  Denote the corresponding set of neighbors as $\cS_j(k)$. The remaining $N_j-p_j(k)$ workers are called the backup workers of $j$ at iteration $k$ and denote them as $\cB_j(k)$.  As a result, each worker first locally updates its parameter at iteration $k$ as in~(\ref{eq:local_update}), and then follows the consensus update only with workers from $\cS_j(k)$ as in~(\ref{eq:consensu_update-partial}).
\beq\label{eq:local_update}
\tilde{\bw}_j(k) &=&\bw_j({k-1})-\eta_j(k) g(\bw_j(k-1)),\\ \label{eq:consensu_update-partial}
\bw_j(k)&=&\sum_{i\in \cS_j(k)}\tilde{\bw}_i(k)P_{i,j}(k),
\eeq
where $\bP(k)=(P_{i,j}(k))$ is the time-varying stochastic consensus matrix at iteration $k$, and $P_{i,j}(k)=0$ if $i\in\cB_j(k)$.
We denote the gradient matrix as $\bG(k)=[g(\bw_1(k)),\ldots,g(\bw_N(k))]\in\R^{d\times N}$ and the learning rate as $\boldsymbol{\eta}(k)=\text{diag}[\eta_1(k),\ldots,\eta_N(k)]\in\R^{N\times N}$.  Then the compact form of \eqref{eq:local_update} and \eqref{eq:consensu_update-partial} satisfies 
\beq
\bW(k)=[\bW({k-1})-\bG(k-1)]\bP(k)\boldsymbol{\eta}(k),
\eeq
from which we have
\beq\nonumber\label{eq:recursion}
\bW(k)=\bW(0)\underset{k}{\underbrace{\bP(1)\cdots\bP(k)}}-\bG(0)\underset{k}{\underbrace{\bP(1)\cdots\bP(k)}} \boldsymbol{\eta}(1) \\
-\bG(1)\underset{k-1}{\underbrace{\bP(2)\cdots\bP(k)}}\boldsymbol{\eta}(2)-\cdots-\bG(k-1)\underset{1}{\underbrace{\bP(k)}}\boldsymbol{\eta}(k).
\eeq
Note that the value of $p_j(k)$, and hence the consensus matrix $\bP(k)$ in our model is not static but dynamically change from one iteration to the other, ensuring faster convergence time.  This will be described in details in Section~\ref{sec:convergence}.

We summarize the workflow of the above consensus-based dynamic backup workers scheme in Algorithm \ref{alg:Framwork}, and call it \texttt{cb-DyBW}.  One key challenge is how to determine the number of dynamic backup workers or equivalently $p_j(k)$ for each worker $\forall j\in[N]$ at each iteration $k$ so as to minimize the globally training (convergence) time over the network.  In the following, we first characterize the convergence performance of \texttt{cb-DyBW} in Section~\ref{sec:convergence} and then determine how to dynamically select the backup workers in Section~\ref{sec:selection}.

\begin{algorithm}
	\caption{Consensus-based Dynamic Backup Workers (\texttt{cb-DyBW})}
	\label{alg:Framwork}
	\begin{algorithmic}[1]
		\Require Network $\cG_0=(\cN,\cE)$, number of iteration $K$.
		\Ensure Estimated parameter $\hat{\bw}$.
		\For{$k=1,2,...,K$}
		\If{$k=1$}
		\State Set $p_j(k)=v_j,j\in[N]$, where $v_j$ is the number of neighbors for worker $j$.
		\EndIf
		\State Compute local gradient and update local parameter as in~(\ref{eq:local_update}) for $\forall j\in[N]$;
		\State Correct local parameter with the received updates from the first $p_j(k)$ neighbors as in~(\ref{eq:consensu_update-partial}) for $\forall j\in[N]$;
		\State $k=k+1;$
		\State Update $p_j({k+1}),$ for $\forall j\in[N]$.
		\EndFor
	\end{algorithmic}
\end{algorithm}

\section{Convergence Analysis}\label{sec:convergence}

In this section, we analyze the convergence of \texttt{cb-DyBW}.  We show that a linear speedup for convergence (i.e., convergence performance increases linearly with respect to the number of workers) is achievable by \texttt{cb-DyBW}.  We relegate all proof details to the appendix.  

\subsection{Assumptions}
We first introduce the assumptions utilized for our convergence analysis.

\begin{assumption}[Non-Negative Metropolis Weight Rule]\label{assumption-weight}
The Metropolis weights on a time-varying graph $(\cN, \cE_k)$ with census matrix $\bP(k)=(P_{i,j}(k))$ at iteration $k$ satisfy 
\beq
\begin{cases}
P_{i,i}(k)= 1-\sum_{j\in\cS_i(k)} P_{i,j}(k),\\
P_{i,j}(k)=\frac{1}{1+\max\{p_i(k), p_j(k)\}}, \quad\text{if $j\in\cS_i(k)$},\\
P_{i,j}(k)=0,\quad \text{otherwise},
\end{cases}
\eeq
where $p_i(k)$ is the number of active neighbors that node $i$ needs to wait at iteration $k.$  Given the Metropolis weights, the matrices $\bP(k)$ are doubly stochastic, i.e., $\sum_{j=1}^N P_{i,j}(k)=\sum_{i=1}^N P_{i,j}(k)=1, \forall i, j, k.$
\end{assumption}

\begin{assumption}[Bounded Connectivity Time]\label{assumption-connectivity}
There exists an integer $B\geq 1$ such that the graph $\cG=(\cN, \cE_{kB} \cup \cE_{kB+1}\cup\ldots\cup \cE_{(k+1)B-1}), \forall k$ is strongly-connected.
\end{assumption}

\begin{assumption}[L-Lipschitz Continuous Gradient]\label{assumption-lipschitz}
We assume that for each worker $j\in[N]$ the loss function $F_j(\cdot): \R^d\rightarrow \R$ is convex and differentiable, and the gradient is Lipschitz continuous with constant $L>0$, i.e.,
\beq
\|\nabla F_j(\bx)-\nabla F_j(\by)\|_2\leq L\|\bx-\by\|, \quad\forall \bx, \by\in\R^d.
\eeq
\end{assumption}

\begin{assumption}[Bounded Variance]\label{assumption-variance}
There exist constants $\sigma_{jL}>0$ for $j\in[N]$ such that the variance of local gradient estimator is bounded by $\mathbb{E}[\|g(\bw_j(k))-\nabla F_j(\bw_j(k))\|^2]\leq \sigma_{jL}^2, \forall j, k.$
\end{assumption}

Assumptions~\ref{assumption-weight}-\ref{assumption-variance} are standard ones in the related literature.  For example, Assumptions~\ref{assumption-weight} and~\ref{assumption-connectivity} are common in dynamic networks with finite number of nodes \cite{Boyd05,tang2018communication}. Specifically, Assumption 1 guarantees the product of consensus matrices $\bP(1)\cdots\bP(k)$ being a doubly stochastic matrix, while Assumption 2 guarantees the product matrix is a strictly positive matrix for large $k$.  We define $\Phi_{k:s}$ as the product of consensus matrices from $\bP(s)$ to $\bP(k)$, i.e., $
\pmb{\Phi}_{k:s}\triangleq \bP(s)\bP(s+1)\cdots \bP(k)$. Under Assumption 2, \cite{nedic2018network} proved that for any integer $\ell, n$, $\pmb{\Phi}_{(\ell+n)B-1:\ell B}$ is a strictly positive matrix.  In fact, every entry of it is at least $\beta^{nB}$, where $\beta$ is the smallest positive value of all consensus matrices, i.e., $\beta=\arg\min_{i, j, k} P_{i,j}(k)$ with $P_{i,j}(k)>0, \forall i, j, k$.

Assumption~\ref{assumption-lipschitz} is widely used in convergence results of gradient methods, e.g., \cite{nedic2009distributed,bubeck2015convex,bottou2018optimization,nedic2018network}.
Assumption~\ref{assumption-variance} is also a standard assumption \cite{kairouz2019advances,tang2020communication}.  We use the bound $\sigma_{jL}$ to quantify the heterogeneity of the non-i.i.d. local datasets across workers.  In particular, $\sigma_L\equiv\sigma_{jL}, \forall j\in[N]$ corresponds to i.i.d. datasets.  For the ease of exposition, we consider a universal bound $\sigma_L$ in our analysis, which can be easily generalized to the non-i.i.d. case.   It is worth noting that we do \textit{not} require a bounded gradient assumption, which is often used in distributed optimization analysis.

\subsection{Main Convergence Results}

\subsubsection{Convergence in terms of Iteration}\label{sec:convergence-iteration}

In this subsection, we analyze the convergence of \texttt{cb-DyBW} in terms of iterations.  We first show that as the number of iteration increases, the gradients tend to be $\pmb{0}$.
\begin{thm}\label{thm:gradient}
Under assumptions~\ref{assumption-weight}-\ref{assumption-variance} and that a constant learning rate $\eta_j(k)=\eta, \forall j, k$ is used, the local gradient of worker $\forall j\in[N]$ generated by \texttt{cb-DyBW} satisfies
\beq
\mathbb{E}[\|\nabla f(\bw_j(k))\|^2]\leq \frac{2 |f(\bw_j(0))-f(\bw^*)|}{\eta k}+\alpha \sigma_L^2, \label{ineq:grad_bound}
\eeq
under a large $k\geq \frac{NB\log_{1-\beta^{NB}}\epsilon+NB+2}{2},$
and $\alpha=4\eta^2 L^3N\frac{(1+\beta^{-NB})^2 (1-\beta^{NB})^{\frac{2k-NB-2}{NB}}}{(1-(1-\beta^{NB})^{-1/NB})^2}+\frac{L\eta}{N}$, $\epsilon>0$ is a constant, with the expectation over the local dataset samples among workers.
\end{thm}
{
\textit{Proof Sketch.} Let
$\by(k)=\frac{1}{N}\sum\limits_{j=1}^{N} [\bw_j(0)-\eta\sum\limits_{s=1}^{k}g(\bw_j(s-1))],$
and~$\by(k+1)=\by(k)-\frac{\eta}{N}\sum\limits_{j=1}^{N} g(\bw_j(k))$.
Since $\nabla f$ is Lipschitz continuous with $L$, we have $\mathbb{E}[f(\by(k+1))]$
\beq\label{eq:expectation_bound}\nonumber
\mathbb{E}[f(\by(k+1))]\leq& f(\by(k))+\mathbb{E}[\nabla f(\by(k))^\intercal(\by(k+1)-\by(k))]
+\frac{L}{2}\mathbb{E}[\|\by(k+1)-\by(k)\|^2]\\ \nonumber
=& \underset{C_1}{\underbrace{\mathbb{E}[\nabla f(\by(k))^\intercal((\by(k+1)-\by(k))+\eta\nabla f(\by(k)))]}}-\eta\|\nabla f(\by(k))\|^2\\
&+f(\by(k))+\frac{L}{2}\underset{C_2}{\underbrace{\mathbb{E}[\|\by(k+1)-\by(k)\|^2]}}.
\eeq
The proof boils down to bound $C_1$ and $C_2$. We show that for sufficient large $k$, $\nabla f(\by(k))$ converges to $\pmb{0}$. The next step is to bound the difference between $\bw_j(k), \forall j$ with $\by(k)$. We show that there exists a $\epsilon>0$ such that $\|\by(k)-\bw_j(k)\|^2\leq \epsilon$ as $k$ is large and above a certain level.  Please see Appendix~\ref{sec:proof-convergence-iteration} for the full proof. 
$\square$
}

\begin{remark}
The bound~\eqref{ineq:grad_bound} on the gradient consists of two parts: (i) ${2|f(\bw_j(0))-f(\bw^*)|}/{\eta k}$ that vanishes as $k$ increases, and (ii) $\alpha \sigma_L^2$ that converges to $\frac{L\eta}{N}\sigma_L^2$ as $k$ increases. This term tends to zero with a large $N$.
\end{remark}

With Theorem~\ref{thm:gradient}, there exists a large enough value of iteration $K$ such that gradients tend to zero after $K$.  This enables us to consider a truncated model in which $\bG(k)=\pmb{0}, \forall k>K.$ Therefore, our model update in \texttt{cb-DyBW}  through~(\ref{eq:recursion}) (or equivalently~(\ref{eq:local_update}) and~(\ref{eq:consensu_update-partial})) reduces to
\beq\label{eq:recursion2}
\bW(k)=\bW(0)\underset{k}{\underbrace{\bP(1)\cdots\bP(k)}}-\bG(1)\underset{k}{\underbrace{\bP(1)\cdots\bP(k)}}\pmb{\eta}(1)
-\cdots-\bG(K)\underset{k-K+1}{\underbrace{\bP(K)\cdots\bP(k)}}\pmb{\eta}(K).
\eeq
It is obvious that \eqref{eq:recursion2} is equivalent to \eqref{eq:recursion} when $k\leq K$. We now analyze the convergence of \texttt{cb-DyBW} under this truncated model, for which we have the following result:

\begin{thm}\label{thm:convergence}
Under assumptions~\ref{assumption-weight}-\ref{assumption-variance} and that a constant learning rate $\eta_j(k)=\eta, \forall j, k$ is used, the sequence of parameters generated by the recursions~\eqref{eq:recursion2} satisfies
\beq\label{eq:convergence_speed}
\mathbb{E}[f(\by(K))]-f(\bw^*)\leq
\frac{\|\by(0)-\bw^*\|^2}{2\eta K}+\frac{L\eta^2}{2N}\sigma_L^2,
\eeq
where $\by(K)=\frac{1}{N}\bW(0)\pmb{1}-\frac{\eta}{N}\sum\limits_{l=1}^{K}\bG(l-1)\pmb{1}$, $\by(0)=\frac{1}{N}\bW(0)\pmb{1}$ and the expectation is over the local dataset samples among workers.
\end{thm}
\textit{Proof Sketch.}
The proof follows a similar approach as that for Theorem~\ref{thm:gradient}.  
It leverages the inequality induced by Lipschitz continuous gradient
\beq\nonumber
&&\hspace{-0.8cm}\mathbb{E}[f(\by_K)]\leq f(\by(K-1))+\underset{C_3}{\underbrace{\frac{1}{2}L\eta^2\mathbb{E}[\|\frac{1}{N}\bG(K-1)\pmb{1}\|^2]}}
+\underset{C_4}{\underbrace{\mathbb{E}[\nabla f(\by(k-1))^\intercal(-\eta\frac{1}{N}\bG(t-1)\pmb{1})]}},
\eeq
and to bound $C_3$ and $C_4$ with respect to (w.r.t.) the loss function $\mathbb{E}[f(\by(K))]$. We show that for large $K$, the expected loss converges to the optimal loss with a linear speedup.  Please see Appendix~\ref{sec:proof-convergence-iteration2} for the full proof. 
$\square$

The following result is an immediate consequence of Theorem~\ref{thm:convergence}. 
{\begin{cor}\label{cor:weight}
Consider the truncated recursions~\eqref{eq:recursion2}, in which $\bW(k)$ denotes the weights at each node. We have
\beq\label{eq:W_converge}
\lim_{k\rightarrow \infty} \bW(k)=\by(K)\pmb{1}^\intercal.
\eeq
\end{cor}}

\begin{remark}
The convergence bound consists of two parts: a vanishing term $\frac{\|\by_0-\bw^*\|^2}{2\eta K}$ as $K$ increases and a constant term $\frac{L\eta^2}{2N}\sigma_L^2$ whose value depends on the problem instance parameters and is independent of $K$.  This convergence bound with dynamic backup worker has the same structure but with different variance terms as that of the typical consensus-based method with full worker participation.  In other words, the decay rate of the vanishing term matches that of the typical consensus-based method.  This implies that the dynamic backup workers does not result in fundamental changes in convergence (in order sense) in terms of iterations.  However, as we will see later, dynamic backup workers can significantly reduce the convergence in terms of wall-clock time since it reduces the length of each iteration.  We further note that this convergence bound can be faster than centralized SGD with less communication on the busiest worker.
\end{remark}

\begin{cor}\label{cor:linear-speedup}
Let $\eta=\sqrt{{N}/{K}}$.  The convergence rate of Algorithm~\ref{alg:Framwork} is $\mathcal{O}\left(\frac{1}{\sqrt{NK}}+\frac{1}{K}\right)$.
\end{cor}
\begin{remark}
The consensus-based distributed optimization methods with dynamic backup workers can still achieve a linear speedup $\mathcal{O}\left(\frac{1}{\sqrt{NK}}\right)$ with proper learning rate settings as shown in Corollary~\ref{cor:linear-speedup}.  Although many works have achieved this convergence rate asymptotically, e.g., \cite{lian2017can} is the first to provide a theoretical analysis of distributed SGD with a convergence rate of $\mathcal{O}(\frac{1}{\sqrt{NK}}+\frac{1}{K})$, these results are only for consensus-based method with full worker participation. It is non-trivial to achieve our result due to the large scale distributed and heterogeneous nature of training data across workers.
\end{remark}

With Corollary~\ref{cor:linear-speedup}, we immediately have the following results on the number of iterations required for convergence:

\begin{cor}\label{cor:convergence-iteration}
Let $\eta=\sqrt{{N}/{K}}$.  The number of iterations required to achieve $\epsilon$-accuracy for the loss function is denoted $\tilde{K}_\epsilon$ and is $\cO\Big(\frac{1}{\epsilon^2N}\Big),$ and the total number of required iterations for the convergence of parameters is $K_\epsilon=\tilde{K}_\epsilon+ \cO(\tilde{K}),$ where $\cO(\tilde{K})$ is the number of iterations to guarantee the convergence of parameter according to \eqref{eq:W_converge} in Corollary 1.
\end{cor}
Note that for $\cO(\tilde{K})$ iterations, there is no local computation of gradients at each worker. They only exchange the information of their parameters to reach a consensus.

\subsubsection{Convergence in terms of Wall-Clock Time}\label{sec:convergence-time}
In this subsection, we analyze the convergence of \texttt{cb-DyBW} in terms of wall-clock time.  We show that dynamic back workers can dramatically reduce the convergence time compared to that with full worker participation.  The intuition is that since each worker only needs to wait for the fastest neighbors to update its local parameter, the length of one iteration can be significantly reduced, which in turn reduces the convergence time given the results on convergence in iterations in Section~\ref{sec:convergence-iteration}.  As a result, we need to estimate the time $T(k)$ needed for each iteration $k.$

We denote the time taken by $j$ to compute its local update at iteration $k$ as $t_j(k)$, which is assumed to be a random variable. W.l.o.g., we assume that each worker consumes different amount of time to compute its local update due to the different sizes of available local training data.  We denote $T_j(k)$ as the time worker $j$ needs to collect updates from $p_j(k)$ neighbors at iteration $k$, which is given by
\beq
T_j(k)=\max\{t_i(k), \forall i\in\cS_j(k)\}.
\eeq
Define $\cV^\prime(k)=\cup_{i=1}^N \cS_i(k)$, which is a subset of $\cN$.
Therefore, the total time $T(k)$ needed for all workers to complete updates at iteration $k$ satisfies
\beq
T(k)\triangleq \max\{ T_j(k), \forall j\in\cV^\prime\}.
\eeq

We approximate $T(k)$ by a deterministic constant such that the mean square error (MSE) is minimized.
The MSE yields from an estimation of $T(k)$ by constant $T$ satisfies
\beq\label{eq:time-mse}
e=\mathbb{E}[(T(k)-T)^2]=\int_0^\infty (T(k)-T)^2 f_{T(k)}(t)dt,
\eeq
where $f_{T(k)}(t)$ is the probability density function (PDF) of $T(k)$.  A necessary condition for the minimization of $e$ can be obtained by taking the derivative of $e$ w.r.t. $T$ and setting it to zero, from which we obtain an MSE estimator as
\beq
T=\int_0^\infty tf_{T(k)}(t)dt=\mathbb{E}[T(k)].
\eeq

With this estimator, we will show that the length of one iteration with dynamic backup workers (i.e., partial worker participation) is smaller than that of conventional consensus-based methods with full worker participation.  We denote the corresponding length of one iteration as $T_p(k)$ and $T_{\text{full}}(t)$, respectively.  We have the following result:

\begin{cor}\label{cor:time_compare}
  The event that $\mathbb{E}[T_{\text{full}}(k)]\geq \mathbb{E}[T_p(k)], \forall k$ occurs almost surely, i.e.,
  \beq
  \mathbb{E}[T_{\text{full}}(k)]\geq \mathbb{E}[T_p(k)], \quad \text{with probability $1$}.
  \eeq
  \end{cor}

\begin{remark}
The expected time for one iteration with dynamic backup workers is smaller than that of full worker participation.  Combined with results in Section~\ref{sec:convergence-iteration} (Theorem~\ref{thm:convergence}), the convergence time of consensus-based distributed optimization methods with dynamic backup workers is reduced compared to that of conventional consensus-based methods.  As we will numerically show in Section~\ref{sec:sim}, our Algorithm~\ref{alg:Framwork} can dramatically reduce the convergence time.
\end{remark}

\section{Dynamic Backup Worker Selection}\label{sec:selection}

Given the convergence result on \texttt{cb-DyBW}, we now discuss how to determine the number of backup workers $b_j(k)$ or $p_j(k)$ for each worker $j\in[N]$ at iteration $k$ for \texttt{cb-DyBW} so that the total convergence time is minimized. 
From Corollary~\ref{cor:convergence-iteration}, the total number of required iterations contains two parts. The first $\tilde{K}_\epsilon$ iterations conduct the stochastic gradient descent, while the second $\cO(\tilde{K})$ iterations are only for the convergence of the product among time-variant consensus matrices. We note that the major time consumption comes from the calculation of gradients, while the communication delay is negligible.  Therefore, we only consider the time for the first $\tilde{K}_\epsilon$ iterations.

\subsection{Distributed Threshold-Based Update Rule (DTUR)}
We propose the following ``threshold based" rule for choosing the set of fastest workers that each worker needs to wait and collect their updated parameters in each iteration.  Within each iteration, each worker waits for a maximum time of $\theta(k)$ for its neighbors to send updates, and then includes them in $\cS_j(k)$. Thus we let $\cS_j(k)=\{i|  i\in\cN_j, t_i(k)\leq \theta(k)\}$ if $t_j(k)\leq \theta(k)$, while $\cS_j(k)=\emptyset$ in case $t_j(k)> \theta(k)$. 
This rule is fully specified by deciding the threshold $\theta(k)$.
~We now analyze its performance under the following optimization problem 
\beq 
\min \sum_{k=1}^{\tilde{K}_\epsilon} \theta(k), \quad \text{s.t. $\tilde{\cG}=\{\cN, \tilde{\cE}\}$ is strong connected}, \label{eq:objective}
\eeq
where $\tilde{\cE}=\cup_{i}^{i+B-1}\cE_i$ for $ i=1,\cdots, T_\epsilon-B+1$.

The optimization problem~(\ref{eq:objective}) is in general hard to solve.  In the following, we relax it and propose a distributed algorithm to solve it.   For a given communication graph $\cG_0=\{\cN, \cE\}$, we first find the shortest path that connects all nodes in this network. Let $\cP$ denote the set of links for the shortest path and $d$ be its length. If there exists more than one shortest path, we randomly select one as $\cP.$   W.l.o.g, we assume the $B$ in \eqref{eq:objective} as $d$, i.e., the dynamic consensus weights matrices are $d$-strongly connected.  Our key insight is that \eqref{eq:objective} can be relaxed to find the minimal time for every $d$ iterations, in which all links in $\cP$ has been visited at least once, i.e., all nodes in $\cP$ share information with each other.  To this end, such a $d$-iteration procedure can run multiple times until the convergence of parameters.  More importantly, every $d$-iteration procedure is independent.

\textbf{A $d$-iteration Procedure.}  
W.l.o.g., we consider one particular $d$-iteration procedure, and we call it an ``epoch" which consists of $d$ iterations.  The algorithm aims to establish at least one connected link $(i, j)$ at each iteration so that $\cP$ is $d$-strongly connected at the end of this epoch.  At the beginning of the $m$-th epoch, denote an empty set $\cP^\prime$ to to store the established links in each iteration during this epoch.  At the end of this epoch, $\cP^\prime$ is reset to be empty.  

For simplicity, we denote the iteration index in the $m$-th epoch as $k$ satisfying $k=md+l$ for $l\in[d].$ At iteration $k$, all workers start their local updates simultaneously.  Once one worker completes its local update, it sends its update to its neighbors and waits for collecting updates from its neighbor as well.  If two workers $i, j$ ($i\in\cN_j$ and $j\in\cN_i$) successfully exchange their local updates, the link $(i, j)$ is established if $(i,j)\in\cP \cap (i,j)\notin\cP^\prime.$  Then link (i, j) is added to $\cP^\prime$, i.e., $\cP^\prime=\cP^\prime\cup \{(i,j)\}$, and all workers move to the $(k+1)$-th iteration.  However, if the established link satisfies 
$(i,j)\notin\cP\cup (i,j)\in\cP^\prime$, then it will not be added to $\cP^\prime$, and iteration $k$ continues until one such link is established.  Therefore, the time for such link establishment is the desired maximum time for one iteration, and we denote it as $\theta(md+l)$, i.e., 
  \beq
\theta(md+\ell)\triangleq\text{min time when $(i,j)\!\in\!\cP$ but $\notin\!\cP^\prime, \forall i, j$}.\label{eq:threshold_time}
 \eeq

Since $\cP^\prime$ is reset to be empty at the end of each independent epoch, we can equivalently minimize the time for each epoch for the optimization problem~(\ref{eq:objective}), i.e., 
\beq 
\min\sum_{\ell=1}^{d} \theta(md+\ell), \quad
\text{s.t.} \cup_{\ell=1}^{d}\cE_{md+l}=\cP.\label{eq:objective_relax}
\eeq
We summarize the workflow for the $d$-iteration procedure in DTUR in  Algorithm~\ref{alg:procedure}. 

\begin{algorithm}
	\caption{A $d$-iteration for Distributed Threshold-Based Update Rule} 
	\label{alg:procedure}
	\begin{algorithmic}[1]
		\Require Network $\cG_0=(\cN,\cE)$.
		\Ensure Find a shortest path  $\cP$ and its length $d$.
		\For{$m=0,1,...,M$}
        \State Set $\cP^\prime=\emptyset$.
        \For{$\ell=1, 2, \ldots, d$}
		\State Find the first link $(i,j)\in\cP$ but $\notin \cP^\prime$,  and store $(i,j)$ into $\cP^\prime$;
        \State Define $\theta(md+\ell)$ according to \eqref{eq:threshold_time};
		\State $\ell=\ell+1;$
        \EndFor
        \State $m=m+1.$
		\EndFor
	\end{algorithmic}
\end{algorithm}

\begin{remark}
Algorithm \ref{alg:procedure} requires each worker to store a local copy of $\cP$, which needs a memory size $\cO(N)$. When a desired link $(i,j)$ is established, workers $i$ and $j$ will broadcast the information $(i,j)$ to the entire network such that each worker can update $\cP^\prime$. In addition, workers $i$ and $j$ need to send a command to the rest workers to terminate the current iteration. The overall communication overhead is $\cO(2Nd)$, which is linear in the number of workers $N$.
\end{remark}

\section{Numerical Results}\label{sec:sim}

In this section, we conduct extensive experiments to validate our model and theoretical results.  For the notation abuse, we use \texttt{cb-DyBW} to denote the algorithm that uses the distributed threshold-based update rule (Algorithm~\ref{alg:procedure}) to determine the number of dynamic backup workers in Algorithm~\ref{alg:Framwork}.  We implement \texttt{cb-DyBW} in Tensorflow \cite{abadi2016tensorflow}  using 
the Network File System (NFS) and MPI backends, where files are shared among workers through MFS and communications between workers are via MPI.  

\begin{figure*}[h]
	\centering
	\includegraphics[width=0.95\textwidth]{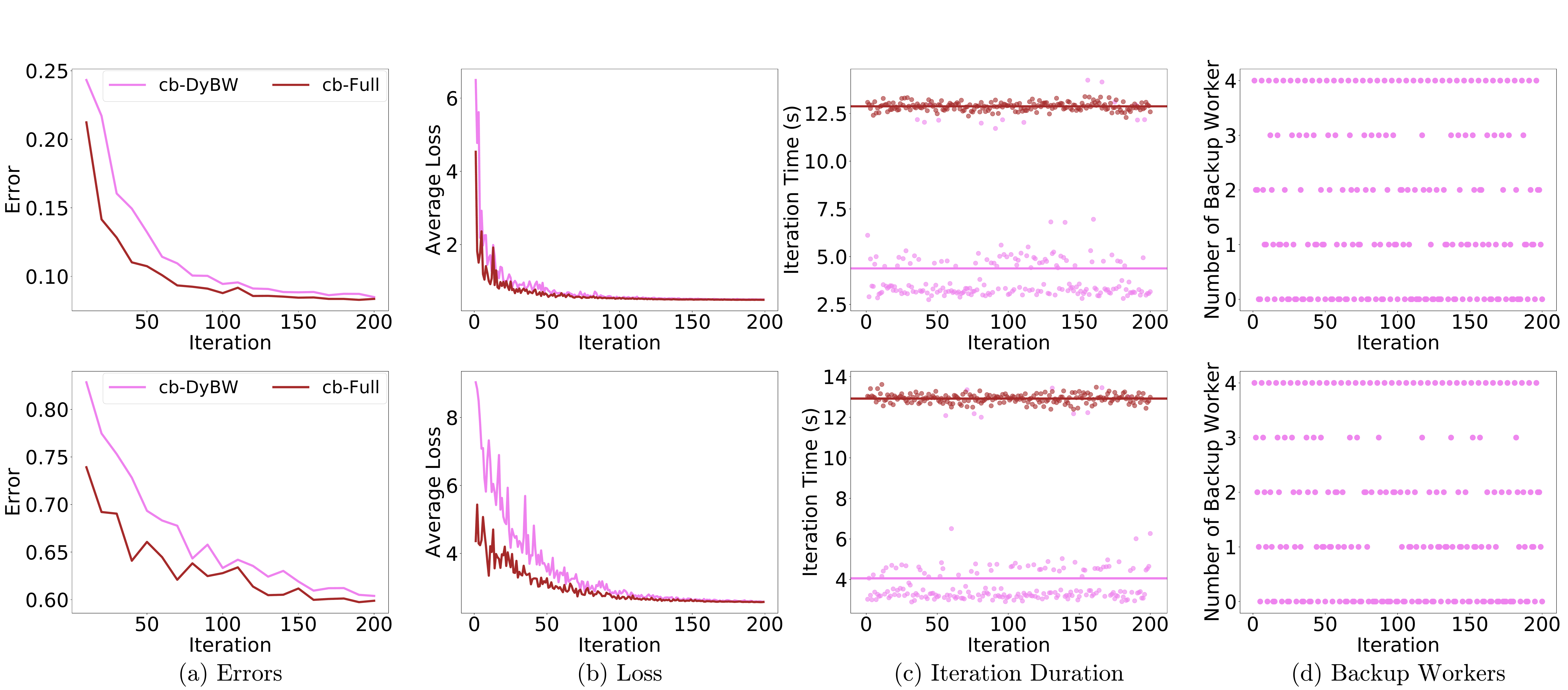}
	\caption{Performance of \texttt{cb-DyBW} and \texttt{cb-Full} for the LRM model under MNIST (top) and CIFAR-10 (bottom).  The straight line in (c) corresponds to the average number of backup workers over the iterations. }
	\label{fig:comparison}
\end{figure*}

We compare \texttt{cb-DyBW} with the conventional consensus-based distributed optimization methods in which full workers participate during the training time.  We call this benchmark as \texttt{cb-Full}.  It is clear that \texttt{cb-Full} suffers the straggler issues in the optimization problem since it ignores the slower workers, which often leads to a longer convergence time compared to that of \texttt{cb-DyBW}.

We evaluate \texttt{cb-DyBW} and \texttt{cb-Full} on the multi-class classification problem.  We use different models including the Logistic Regression Model (LRM) and a fully-connected neural network with 2 hidden layers (2NN) with MNIST \cite{lecun1998gradient,minst} and CIFAR-10 \cite{krizhevsky2009learning,cifar} datasets.  The MNIST dataset contains handwritten digits with $60,000$ samples for training and $10,000$ samples for testing.  The CIFAR-10 dataset consists of $60,000$ $32\times32$ color images in $10$ classes where $50,000$ samples are for training and the other $10,000$ samples for testing.  We consider a network with $6$ workers and randomly generate a connected graph for evaluation.  The loss function we consider is the cross-entropy one.  For the ease of exposition, we relegate some experimental results including testing different models, and the impact of different network topologies with different number of workers etc to Appendix~\ref{app:sim}.  

To enhance the training efficiency, we reduce the dimensions of MNIST (the dimension of samples is $784$) and CIFAR-10 (the dimension of samples is $3,072$) through the widely used principal component analysis (PCA) \cite{wold1987principal}.   
The learning rate is perhaps the most critical hyperparameter in distributed ML optimization problems.  A proper value of the learning rate is important; however, it is in general hard to find the optimal value.  The standard recommendation is to have the learning rate proportional to the aggregate batch size with full worker participation \cite{goyal2017accurate,smith2018don}.  With the consideration of dynamic backup workers, it is reasonable to adaptively set the learning rate \cite{chen2016revisiting,xu2020dynamic}.  Hence, we choose $\eta(k)=\eta_0\cdot\delta^k$ where $\eta_0=0.2$ and $\delta=0.95.$  Finally, batch size is another important hyperparameter, which is limited by the memory and computational resources available at each worker, or determined by generalization performance of the final model \cite{hoffer2017train}.   We test the impact of batch size using these two datasets and find that $1,024$ is a proper value.  For the ease of exposition, we relegate the detailed comparisons to Appendix~\ref{app:sim}.

Figure~\ref{fig:comparison} shows the (testing) errors, (training) loss and iteration duration of \texttt{cb-DyBW} and \texttt{cb-Full} for the LRM model under MNIST (top) and CIFAR-10 (bottom) datasets, as well as the number of dynamic backup workers for \texttt{cb-DyBW}.  We observe that the number of iterations required for convergence is similar (in order sense) for both \texttt{cb-DyBW} and \texttt{cb-Full}, consistent with our theoretical results in Theorem \ref{thm:convergence}.  However, it is clear that \texttt{cb-DyBW} can dramatically reduce the duration of one iteration by 65\%-70\% on average compared to that of \texttt{cb-Full} from Figure~\ref{fig:comparison} (c).   This is because our proposed framework and algorithm \texttt{cb-DyBW} can dynamically and adaptively determine the number of backup workers for each worker during the training time so as to mitigate the effect of stragglers in the optimization problem.  As a result, \texttt{cb-DyBW} can significantly reduce the convergence time compared to \texttt{cb-Full} for a certain accuracy.  Finally, from Figure~\ref{fig:comparison} (d), it is clear that the number of backup workers is dynamically changing over time during the training time, which further validates our motivation and model.

\section{Conclusions}\label{sec:con}

In this paper, we considered to mitigate the effect of stragglers in distributed machine learning via dynamic backup workers.  We formulated the consensus-based distributed optimization problem with dynamic backup workers to minimize the total training time for a certain convergence accuracy.  We proposed \texttt{cb-DyBW} and analyzed its convergence.  We proved that \texttt{cb-DyBW} achieves a linear speedup for convergence, and can dramatically reduce the convergence time compared to conventional consensus-based methods.  We further proposed a threshold-based rule to determine how to dynamically select backup workers during the training time.  Finally, we provided empirical experiments to validate our theoretical results.

\bibliographystyle{unsrt}  
\bibliography{refs}

\clearpage
\appendix

\section{Proofs of Main Results}

In this section, we provide the proofs of the theoretical results presented in the paper.  We first prove that the gradients tends to be $\pmb{0}$ after a large number of iterations for \texttt{cb-DyBW} (Theorem~\ref{thm:gradient}), and then provide proofs for the convergence of \texttt{cb-DyBW} (Theorem~\ref{thm:convergence}).

\subsection{Proof of Theorem~\ref{thm:gradient}}\label{sec:proof-convergence-iteration}
\begin{proof}
 We define $\pmb{\Phi}_{k:s}$ as the product of consensus matrices from $\bP(s)$ to $\bP(k)$, i.e.,
$\pmb{\Phi}_{k:s}\triangleq \bP(s)\bP(s+1)\cdots \bP(k),$
and let ${\Phi}_{k,s}(i,j)$ be the $i$-th row $j$-th column element of $\pmb{\Phi}(k:s)$.
Therefore, with fixed $\eta$, $\bw_i(k)$ can be expressed in terms of $\pmb{\Phi}$ as
\beq
\bw_i(k+1)=\sum\limits_{j=1}^{N}\Phi_{k,1}(i,j)\bw_j(0)
-\eta\sum\limits_{r=1}^{k}\sum\limits_{j=1}^{N}\Phi_{k,r}(i,j) g(\bw_j(r-1)).
\eeq

Next, we define an auxiliary variable $\by(k)$ satisfying
\beq
\by(k)=\frac{1}{N}\sum\limits_{j=1}^{N}\bw_j(0)-\eta\sum\limits_{r=1}^{k}\sum\limits_{j=1}^{N}\frac{1}{N} g(\bw_j(r-1)),
\eeq
with the following relation holds
\beq\label{eq:y_recursion}
\by(k+1)=\by(k)-\frac{\eta}{N}\sum\limits_{j=1}^{N} g(\bw_j(k)).
\eeq
From Assumption~\ref{assumption-lipschitz}, we have the following inequality with respect to $\by(k)$ \cite{bottou2018optimization}
\beq\label{eq:expectation_bound}\nonumber
\mathbb{E}\big[f(\by(k+1))\big]\!\!\!\!\!\!&\leq&\!\!\!\!\! f(\by(k))+\mathbb{E}\left[\nabla f(\by(k))^\intercal(\by(k+1)-\by(k))\right]+\frac{L}{2}\mathbb{E}\left[\|\by(k+1)-\by(k)\|^2\right]\\\nonumber
&= &\!\!\!\!\!f(\by(k))-\eta\|\nabla f(\by(k))\|^2+\underset{C_1}{\underbrace{\mathbb{E}\left[\nabla f(\by(k))^\intercal((\by(k+1)-\by(k))+\eta\nabla f(\by(k)))\right]}}\\
&&\qquad\qquad+\underset{C_2}{\underbrace{\frac{L}{2}\mathbb{E}\left[\|\by(k+1)-\by(k)\|^2\right]}}.
\eeq
As a result, we need to bound $C_1$ and $C_2$ in \eqref{eq:expectation_bound}.  Note that the term $C_1$ can be bounded as follows
\beq \nonumber
C_1&= &\!\!\!\!\Bigg \langle  \nabla f(\by(k)), \mathbb{E}\left[(\by(k+1)-\by(k))+\eta\nabla f(\by(k))\right]\Bigg \rangle\\ \nonumber
&\overset{(a_1)}{=}&\!\!\!\!\left \langle \nabla f(\by(k)), \mathbb{E}\left[-\frac{\eta}{N}\sum\limits_{j=1}^{N} g(\bw_j(k))+\frac{\eta}{N}\sum\limits_{j=1}^{N} \nabla F_j(\by(k))\right]  \right \rangle \\ \nonumber
&\overset{(a_2)}{=}&\!\!\!\!\left \langle \sqrt{\eta}\nabla f(\by(k)), \mathbb{E}\left[-\frac{\sqrt{\eta}}{N}\sum\limits_{j=1}^{N}( g(\bw_j(k))-\nabla F_j(\by(k)))\right]  \right \rangle\\ \nonumber
&\overset{(a_3)}{=}&\!\!\!\!\!\frac{\eta}{2}\left\|\nabla f(\by(k))\right\|^2\!+\!\frac{\eta}{2N^2}\!\left\|\mathbb{E}\!\sum\limits_{j=1}^{N}(g(\bw_j(k))\!-\!\nabla F_j(\by(k)))\right\|^2\!\!\!-\!\!\frac{\eta}{2N^2}\!\!\left\|\mathbb{E}\!\sum\limits_{j=1}^{N} g(\bw_j(k))\right\|^2\\
&\overset{(a_4)}{\leq}&\!\!\!\!\! \frac{\eta}{2}\left\|\nabla f(\by(k))\right\|^2\!+\!\underset{D_1}{\underbrace{\frac{\eta}{2N}\!\!\sum\limits_{j=1}^{N}\left\|\mathbb{E}[ g(\bw_j(k))\!-\!\nabla F_j(\by(k))]\right\|^2}}\!\!-\!\!\frac{\eta}{2N^2}\!\!\left\|\mathbb{E}\!\sum\limits_{j=1}^{N} g(\bw_j(k))\right\|^2\!\!\!,
\eeq
where $(a_1)$ follows from \eqref{eq:y_recursion} and the fact that $\frac{1}{N}\sum\limits_{j=1}^{N} \nabla F_j(\by(k))$ is an unbiased estimator of $\nabla f(\by(k)).$ $(a_2)$ comes from a standard mathematical manipulation.  The equality in $(a_3)$ is based on the equation $\langle \ba, \bb \rangle=\frac{1}{2}\left[\|\ba\|^2+\|\bb\|^2-\|\ba-\bb\|^2\right].$ The last inequality in $(a_4)$ is due to the well triangle-inequality $\left\|\sum\limits_{i=1}^N \ba_i\right\|^2\leq N\sum\limits_{i}^N \left\|\ba_i\right\|^2.$

Next, we bound $D_1$ as follows
\beq\nonumber
D_1&\overset{(b_1)}{\leq}&\!\!\!\frac{\eta L^2}{2N}\sum\limits_{j=1}^{N}\|\mathbb{E}[ \bw_j(k)-\by(k)]\|^2\\ \nonumber
&\overset{(b_2)}{\leq}&\!\!\!\frac{\eta L^2}{2N}\sum\limits_{j=1}^{N}\mathbb{E}\left[\|( \bw_j(k)-\by(k))\|^2\right]\\ \nonumber
&=&\!\!\!\frac{\eta L^2}{2N}\sum\limits_{j=1}^{N}\mathbb{E}\Bigg[\Bigg\|\frac{1}{N}\sum\limits_{i=1}^{N} \bw_i(0)-\frac{\eta}{N}\sum\limits_{s=1}^{k}\sum\limits_{i=1}^{N}g(\bw_i(s-1))-\sum\limits_{i=1}^{N}\bw_i(0)\Phi_{k:1}(i,j)\\ \nonumber
&&\qquad\qquad\qquad\qquad+\eta\sum\limits_{s=1}^{k}\sum\limits_{i=1}^{N}g(\bw_i(s-1))\Phi_{k:s}(i,j) \Bigg\|^2\Bigg]\\ \nonumber
&=&\!\!\!\frac{\eta L^2}{2N}\sum\limits_{j=1}^{N}\mathbb{E}\left[\left\|\sum\limits_{i=1}^{N} \bw_i(0)(\frac{1}{N}-\Phi_{k:1}(i,j))-\eta\sum\limits_{s=1}^{k}\sum\limits_{i=1}^{N}g(\bw_i(s-1))(\frac{1}{N}-\Phi_{k:s}(i,j)) \right\|^2\right]\\ \nonumber
&\overset{(b_3)}{\leq}&\!\!\!\frac{\eta L^2}{2N}\sum\limits_{j=1}^{N}\mathbb{E}\Bigg[\left\|\sum\limits_{i=1}^{N} \bw_i(0)(\frac{1}{N}-\Phi_{k:1}(i,j))\right\|^2\\ \nonumber
&&\qquad\qquad+\left\|\eta\sum\limits_{s=1}^{k}\sum\limits_{i=1}^{N}g(\bw_i(s-1))(\frac{1}{N}-\Phi_{k:s}(i,j)) \right\|^2\Bigg]\\ \nonumber
&\overset{(b_4)}{\leq}&\!\!\!\frac{\eta L^2}{2N}\sum\limits_{j=1}^{N}\mathbb{E}\left[\left\|2\eta\sum\limits_{s=1}^{k}\sum\limits_{i=1}^{N}g(\bw_i(s-1))\frac{1+\beta^{-NB}}{1-\beta^{NB}}(1-\beta^{NB})^{(k-s)/NB} \right\|^2\right]\\ \nonumber
&\overset{(b_5)}{=}&\!\!\!\frac{\eta L^2}{2N}\sum\limits_{j=1}^{N}\left\|\mathbb{E}\left[2\eta\sum\limits_{s=1}^{k}\sum\limits_{i=1}^{N}g(\bw_i(s-1))\frac{1+\beta^{-NB}}{1-\beta^{NB}}(1-\beta^{NB})^{(k-s)/NB} \right]\right\|^2\\ \nonumber
&&\qquad\qquad+\frac{\eta L^2}{2N}\sum\limits_{j=1}^{N}\Var\left[\left\|2\eta\sum\limits_{s=1}^{k}\sum\limits_{i=1}^{N}g(\bw_i(s-1))\frac{1+\beta^{-NB}}{1-\beta^{NB}}(1-\beta^{NB})^{(k-s)/NB} \right\|\right]\\ \nonumber
&\overset{(b_6)}{=}&\!\!\!2\eta^3L^2\left\|\mathbb{E}\left[\sum\limits_{s=1}^{k}\sum\limits_{i=1}^{N}g(\bw_i(s-1))\frac{1+\beta^{-NB}}{1-\beta^{NB}}(1-\beta^{NB})^{(k-s)/NB} \right]\right\|^2\\ \nonumber
&&\qquad\qquad+2\eta^3L^2\Var\left[\left\|\sum\limits_{s=1}^{k}\sum\limits_{i=1}^{N}g(\bw_i(s-1))\frac{1+\beta^{-NB}}{1-\beta^{NB}}(1-\beta^{NB})^{(k-s)/NB} \right\|\right]\\ \nonumber
&\overset{(b_7)}{\leq}&\!\!\!2\eta^3L^2\left(\frac{1+\beta^{-NB}}{1-\beta^{NB}}\frac{(1-\beta^{NB})^{(k-1)/NB}}{(1-(1-\beta ^{NB})^{-1/B_0})^2} \right)^2\\\nonumber
&&\qquad\qquad\cdot\left(\left\|\mathbb{E}\left[ \max_s\sum\limits_{i=1}^{N}g(\bw_i(s-1)) \right]\right\|^2+\Var\left[\left\|\max_s\sum\limits_{i=1}^{N}g(\bw_i(s-1)) \right\|\right]\right)\\ \nonumber
&\overset{(b_8)}{\leq}&\!\!\!2\eta^3L^2\left(\frac{1+\beta^{-NB}}{1-\beta^{NB}}\frac{(1-\beta^{NB})^{(k-1)/NB}}{(1-(1-\beta ^{NB})^{-1/B_0})^2} \right)^2\\
&&\qquad\qquad\cdot\left(\left\|\mathbb{E}\left[ \max_s\sum\limits_{i=1}^{N}g(\bw_i(s-1)) \right]\right\|^2+N\sigma_L^2 \right)\label{eq:D1}.
\eeq
The inequality in $(b_1)$ directly comes from Assumption~\ref{assumption-lipschitz} and $(b_2)$ is based on the fact that $\mathbb{E}[\bX]^2\leq \mathbb{E}[\bX^2]$. $(b_3)$ is due to the inequality $\|\ba-\bb\|^2\leq \|ba\|^2+\|\bb\|^2$. Without loss of generality, we assume the initial term $\bw_i(0), \forall i$ is small enough and we can neglect it. Hence, $(b_4)$ holds according to Lemma \ref{lemma_bound_Phi} (see Section~\ref{sec:app-auxiliary}).  $(b_5)$ follows $\mathbb{E}[\bX^2]=\Var[\bX]+\mathbb{E}[\bX]^2$ and $(b_6)$ leverages the property that the workers are independent with each other. $(b_7)$ is the standard mathematical manipulation and $(b_8)$ holds due to the bounded variance in Assumption~\ref{assumption-variance}.

Substituting $D_1$ into $C_1$ yields
\beq \nonumber
C_1\!\!\!\!\!&=&\!\!\!\!\!\frac{\eta}{2}\left\|\nabla f(\by(k))\right\|^2\!-\!\!\frac{\eta}{2N^2}\left\|\mathbb{E}\sum\limits_{j=1}^{N} g(\bw_j(k))\right\|^2\!\!+\!2\eta^3L^2\left(\frac{1+\beta^{-NB}}{1-\beta^{NB}}\frac{(1-\beta^{NB})^{(k-1)/NB}}{(1-(1-\beta ^{NB})^{-1/B_0})^2} \right)^2\\
&&\qquad\qquad\cdot\left(\left\|\mathbb{E}\left[ \max_s\sum\limits_{i=1}^{N}g(\bw_i(s-1)) \right]\right\|^2+N\sigma_L^2 \right).
\eeq

Next, we bound $C_2$ as
\beq\nonumber
C_2\!\!\!&=&\!\!\!\frac{\eta^2L}{2N^2}\mathbb{E}\left[\left\|\sum\limits_{j=1}^{N}g(\bw_j(k))\right\|^2\right]\\\nonumber
&\!\!\!\overset{(c_1)}{=}&\!\!\!\frac{\eta^2L}{2N^2}\left(\mathbb{E}\left[\left\|\sum\limits_{j=1}^{N}g(\bw_j(k))-\nabla  F_j(\bw_j(k)\right\|^2\right]+\left\|\mathbb{E}\sum\limits_{j=1}^{N}g(\bw_j(k))\right\|^2\right)\\
&\!\!\!\overset{(c_2)}{\leq}&\!\!\! \frac{\eta^2L}{2N}\sigma^2_L+\frac{\eta^2L}{2N^2}\left\|\mathbb{E}\sum\limits_{j=1}^{N}g(\bw_j(k))\right\|^2,
\eeq
where $(c_1)$ is due to $\mathbb{E}[\bX^2]=\Var[\bX]+\mathbb{E}[\bX]^2$ and $(c_2)$ follows the bounded variance in Assumption~\ref{assumption-variance}.

Substituting $C_1$ and $C_2$ into the inequality \eqref{eq:expectation_bound}, we have
\beq \nonumber
\mathbb{E}[f(\by(k+1))] \!\!\!&\leq&\!\!\! f(\by(k))-\eta \|\nabla f(\by(k))\|^2+\frac{\eta}{2}\|\nabla f(\by(k))\|^2+\frac{\eta^2L}{2N}\sigma^2_L\\\nonumber
&&+\frac{\eta^2L}{2N^2}\left\|\mathbb{E}\sum\limits_{j=1}^{N}g(\bw_j(k))\right\|^2-\frac{\eta}{2N^2}\left\|\mathbb{E}\sum\limits_{j=1}^{N} g(\bw_j(k))\right\|^2\\ \nonumber
&&+2\eta^3L^2\left(\frac{1+\beta^{-NB}}{1-\beta^{NB}}\frac{(1-\beta^{NB})^{(k-1)/NB}}{(1-(1-\beta ^{NB})^{-1/B_0})^2} \right)^2\\\nonumber
&&\cdot\left(\left\|\mathbb{E}\left[ \max_s\sum\limits_{i=1}^{N}g(\bw_i(s-1)) \right]\right\|^2+N\sigma_L^2 \right)\\\nonumber
&\!\!\!\leq&\!\!\! f(\by(k))-\frac{\eta}{2}\|\nabla f(\by(k))\|^2 \\
&&+ 2\eta^3L^2\left(\frac{1+\beta^{-NB}}{1-\beta^{NB}}\frac{(1-\beta^{NB})^{(k-1)/NB}}{(1-(1-\beta ^{NB})^{-1/B_0})^2} \right)^2N\sigma_L^2+\frac{L \eta^2}{2N}\sigma^2_L. \label{eq:gradiengt_iteration}
\eeq
{The $\mathbb{E}[f(\by(k))]$  in \eqref{eq:gradiengt_iteration} is monotonically decreasing if $\eta\leq\frac{\sqrt{\frac{N\|\nabla f(\by(k))\|^2}{\sigma_L^2}+\frac{1}{16\gamma^2N^2}}-\frac{1}{4\gamma N}}{2LN\gamma}$, where $\gamma=\frac{1+\beta^{-NB}}{1-\beta^{NB}}\frac{(1-\beta^{NB})^{(k-1)/NB}}{(1-(1-\beta ^{NB})^{-1/B_0})^2}.$}
Rearranging the order of each items in \eqref{eq:gradiengt_iteration}, we have
\beq\label{eq:gradient_2} \nonumber
\frac{\eta}{2}\|\nabla f(\by(k))\|^2\!\!\!&\leq&\!\!\! \mathbb{E}[f(\by(k))]-\mathbb{E}[f(\by(k+1))]+\frac{L \eta^2}{2N}\sigma^2_L\\
&&\quad\qquad+ 2\eta^3L^2\left(\frac{1+\beta^{-NB}}{1-\beta^{NB}}\frac{(1-\beta^{NB})^{(k-1)/NB}}{(1-(1-\beta ^{NB})^{-1/B_0})^2} \right)^2N\sigma_L^2.
\eeq
Summing the recursion in \eqref{eq:gradient_2} from the $0$-th iteration to the $k$-th iteration yields
\beq
\sum\limits_{\ell=0}^k \frac{\eta}{2}\|\nabla f(\by(\ell))\|^2&\!\!\!\leq&\!\!\! \mathbb{E}[f(\by(0))]-\mathbb{E}[f(\by(k+1))]+\frac{(k+1)L \eta^2}{2N}\sigma^2_L\\
&&+ 2(k+1)\eta^3L^2\left(\frac{1+\beta^{-NB}}{1-\beta^{NB}}\frac{(1-\beta^{NB})^{(k-1)/NB}}{(1-(1-\beta ^{NB})^{-1/B_0})^2} \right)^2N\sigma_L^2.
\eeq
Due to the fact that the loss function $f$ is convex and $\mathbb{E}[f(\by(k))]$ is non-increasing, we have $\|\nabla f(\by(\ell))\|^2$ being non-increasing.
Define
$$\alpha\triangleq 4\eta^2 L^3N\frac{(1+\beta^{-NB})^2 (1-\beta^{NB})^{\frac{2k-NB-2}{B_0}}}{(1-(1-\beta^{NB})^{-1/NB})^2}+\frac{L\eta}{N}.$$
 Therefore, we achieve
\beq \nonumber
\|\nabla f(\by(k))\|^2
&\!\!\!\leq&\!\!\!\frac{2\left|\mathbb{E}[f(\by(0))]-\mathbb{E}[f(\by(k+1))]\right|}{k\eta}+\alpha\sigma^2_L\\
&\!\!\!\leq& \!\!\!\frac{2 |f(\by(0))-f(\bw^*)|}{k\eta}+\alpha\sigma^2_L, \label{eq:convergence1}
\eeq
where the second inequality holds due to the fact that $f(\bw^*)\leq \mathbb{E}[f(\by(k+1))]$.
According to $(D_1)$ in \eqref{eq:D1}, we have
\beq\nonumber
\mathbb{E}[\bw_j(k)-\by(k)]\!\!\! &\leq&\!\!\! 4\eta^2\Bigg(\frac{1+\beta^{-NB}}{(1-(1-\beta^{NB})^{-1/B_0})^2}\Bigg)^2(1-\beta^{NB})^{(2k-2-NB)/NB}\\
&&\qquad\cdot\left(\left\|\mathbb{E}\left[ \max_s\sum\limits_{i=1}^{N}g(\bw_i(s-1)) \right]\right\|^2+N\sigma_L^2 \right), \forall j,
\eeq
Therefore, there exists an arbitral small constant $\epsilon>0$ and $\delta>0$ such that
\beq
\mathbb{E}[\bw_j(k)-\by(k)]\leq \delta,\quad (1-\beta^{NB})^{(2k-2-NB)/NB}\leq \epsilon,
\eeq
hence
\beq
\mathbb{E}[\bw_j(k)-\by(k)]\rightarrow 0, \quad\forall k\geq \frac{NB\log_{1-\beta^{NB}}\epsilon+NB+2}{2}.
\eeq
Replacing $\by(k)$ with $\bw_j(k)$ in \eqref{eq:convergence1} leads to the desired results. This completes the proof.
\end{proof}

\subsection{Proof of Theorem~\ref{thm:convergence}}\label{sec:proof-convergence-iteration2}

\begin{proof}
This proof contains two parts, i.e., convergence of parameters $\bw_j(k), \forall j$ and convergence of the loss function.

\textit{Convergence of Parameters.}  We first characterize the convergence of parameters.  According to Lemma \ref{themorem_convergence_Phi} (see Section~\ref{sec:app-auxiliary}), the following equation holds for $\bW(k), \forall k>K$,
\beq\label{eq:limit_truncated_recursion}
\lim_{k\rightarrow \infty} \bW(k)=\frac{1}{N}\bW(0)\pmb{1}\pmb{1}^T-\frac{1}{N}\sum\limits_{l=1}^{K}\eta_i\bG(l-1)\pmb{1}\pmb{1}^T,
\eeq
which is irrelevant with $k$ being large. In this truncated model, we know for sure that $\bW(k+1)=\bW(k)$ when $k$ is large, and each column of $\bW_k$ is identical, i.e., $\bw_1(k)=\bw_2(k)=\ldots=\bw_N(k)$. These imply the convergence of the parameters.

\textit{Convergence of Loss Function.}  We now show the convergence of loss function.
In the following we show that the converged parameter in \eqref{eq:limit_truncated_recursion} guarantees the convergence of our loss function.
We denote the r.h.s of \eqref{eq:limit_truncated_recursion} with a fixed step size $\eta$ as
\beq\label{eq:y_definition}
\by(t)=\frac{1}{N}\bW(0)\pmb{1}-\frac{\eta}{N}\sum\limits_{i=1}^{t}\bG(i-1)\pmb{1},\quad \forall t\leq K.
\eeq
Hence, $\by(t)=\by(t-1)-\frac{\eta}{N}\bG(t-1)\pmb{1}$, $\forall t\leq K$ holds.
Since $\nabla f$ is Lipschitz continuous with $L$ according to Assumption~\ref{assumption-lipschitz}, we have the following inequality
\beq\label{eq:inequality_t_t-1}\nonumber
\mathbb{E}[f(\by(t))]\!\!\!\!\!&\leq&\!\!\!\!\! f(\by(t-1))\!+\!\mathbb{E}\left[\nabla f(\by(t-1))^\intercal\left(-\eta\frac{1}{N}\bG(t-1)\pmb{1}\right)\right]\!+\!\frac{1}{2}L\eta^2\mathbb{E}\left[\left\|\frac{1}{N}\bG(t-1)\pmb{1}\right\|^2\right]\\\nonumber
&\!\!\!\!\!=&\!\!\!\!\!f(\by(t-1))-\eta\|\nabla f(\by(t-1))\|^2\\\nonumber
&&\qquad\qquad+\frac{1}{2}L\eta^2\left(\|\nabla f(\by(t-1))\|^2+\frac{1}{N^2}\Var[\|\bG(t-1)\pmb{1}\|]\right)\\
&\!\!\!\!\!\leq&\!\!\!\!\! f(\by_{t-1})-\eta\left(1-\frac{1}{2}L\eta\right)\|\nabla f(\by(t-1))\|^2+\frac{L\eta^2}{2N}\sigma_L^2,
\eeq
where $\mathbb{E}[f(\by_t)]$ is monotonically decreasing as $t$ when $\eta\leq\min\left(\frac{2\|\nabla f(\by(t-1))\|^2}{L(\|\nabla f(\by(t-1))\|^2+\sigma^2_L/N)}, 1/L\right)$.

Due to convexity of $f(\cdot)$, we have the following inequality
$$
f(\by(t-1))\leq f(\bw^*)+\nabla f(\by(t-1))^T(\by(t-1)-\bw^*),
$$
where $f(\bw^*)$ is the optimal value of loss function.
Replacing $f(\by(t-1))$ in \eqref{eq:inequality_t_t-1} by its upper bound, and therefore we have the following
\beq\label{eq:difference_iteration}
\mathbb{E}[f(\by(t))]-f(\bw^*)\leq \frac{1}{4\eta(1-L\eta/2)}\left(\mathbb{E}[\|\by(t-1)-\bw^*\|^2]-\mathbb{E}[\|\by(t)-\bw^*\|^2]\right)+\frac{L\eta^2}{2N}\sigma_L^2.
\eeq
Summing \eqref{eq:difference_iteration} over iterations, we have
\beq\nonumber
\sum\limits_{t=1}^{K}\mathbb{E}[f(\by(t))]-f(\bw^*)&\leq& \frac{1}{4\eta(1-L\eta/2)}(\|\by(0)-\bw^*\|^2-\mathbb{E}[\|\by(k)-\bw^*\|^2])+\frac{KL\eta^2}{2N}\sigma_L^2\\\nonumber
&\leq& \frac{1}{4\eta(1-L\eta/2)}(\|\by(0)-\bw^*\|^2)+\frac{KL\eta^2}{2N}\sigma_L^2\\
&\leq& \frac{1}{2\eta}(\|\by(0)-\bw^*\|^2)+\frac{KL\eta^2}{2N}\sigma_L^2.
\eeq
Since $\mathbb{E}[f(\by(t))]$ is a decreasing function over $t$, we finally conclude that
\beq
\mathbb{E}[f(\by(K))]-f(\bw^*)\leq \frac{1}{2\eta K}(\|\by(0)-\bw^*\|^2)+\frac{L\eta^2}{2N}\sigma_L^2.
\eeq
\end{proof}
This completes the proof.

\subsection{Proof of Corollaries}

\textbf{Proof of Corollary~\ref{cor:weight}.}
\begin{proof}
This directly results from Lemma \ref{themorem_convergence_Phi} and definition of $\by(K)$ in \eqref{eq:y_definition}. 
\end{proof}

\textbf{Proof of Corollary~\ref{cor:linear-speedup}.}
\begin{proof}
 The desired results is yield by substituting $\eta=\sqrt{N/K}$ back in to \eqref{eq:convergence_speed} in Theorem \ref{thm:convergence}. 
\end{proof}

\textbf{Proof of Corollary~\ref{cor:convergence-iteration}}
\begin{proof}
Substituting $\eta=\sqrt{N/K}$ into \eqref{eq:convergence_speed} yields the required number of iterations for achieving $\epsilon$-accuracy as $\tilde{K}_\epsilon=\cO\Big(\frac{1}{\epsilon^2N}\Big).$
Based on the recursion in \eqref{eq:limit_truncated_recursion}, an additional $\cO(\tilde{K})$ iterations are needed for the convergence of parameters. This completes the proof.
\end{proof}

\textbf{Proof of Corollary~\ref{cor:time_compare}}
\begin{proof}Denote the time for full worker participant and partial worker participant at iteration $k$ as $T_{\text{full}}(k)$ and $T_p(k)$, respectively.
The cumulative density function (CDF) of $T_{\text{full}}(k)$ and $T_p(k)$is given by
\beq\label{eq:time-iteration-cdf}\nonumber
\mathbb{P}(T_{\text{full}}(k)<x)&=&\mathbb{P}(\max\{t_j(k), \forall j\in\{1,\ldots,N\}\}<x)\\ \nonumber
&=&\mathbb{P}\Bigg(\prod_{i=1}^{N} t_i(k)<x\Bigg)\\
&=&\prod_{i=1}^N\mathbb{P}(t_i(k)<x),
\eeq
and
\beq\label{eq:time-iteration-cdf}\nonumber
\mathbb{P}(T_p(k)<x)&=&\mathbb{P}(\max\{T_j(k), \forall j\in\cV^\prime(k)\}<x)\\ \nonumber
&=&\mathbb{P}\Bigg(\prod_{i\in \cS_{\tau}(k)} t_i(k)<x\Bigg)\\
&=&\prod_{i\in \cS_{\tau}(k)}\mathbb{P}(t_i(k)<x),
\eeq
respectively, where $\tau=\arg\max_j T_j(k)$. Therefore, the expectation of $T_{\text{full}}(k)$ and $T_p(k)$ are given by
\beq
\mathbb{E}[T_{\text{full}}(k)]=\int_0^\infty 1-\prod_{i=1}^N\mathbb{P}(t_i(k)<x)dx,
\eeq
and
\beq
\mathbb{E}[T_p(k)]=\int_0^\infty 1-\prod_{i\in \cS_{\tau}(k)}\!\!\mathbb{P}(t_i(k)<x)dx,
\eeq
respectively.  Because $t_i(k), \forall i$ follows the same distribution,
\beq
\prod_{i=1}^N\mathbb{P}(t_i(k)<x)\leq \prod_{i\in \cS_{\tau}(k)}\!\!\mathbb{P}(t_i(k)<x), \forall x, \text{with probability 1}.
\eeq
This completes the proof.

\end{proof}

\subsection{Auxiliary Lemmas}\label{sec:app-auxiliary}
We provide the following auxiliary lemmas which are used in our proofs.  We omit the proofs of these lemmas for the ease of exposition and refer interested readers to  \cite{Boyd05} and  \cite{nedic2009distributed} for details. 

\begin{lem}[Theorem 2 in \cite{Boyd05}]
Assume that $\bP(k)$ is doubly stochastic for all $k$. We denoted the limit matrix of  $\Phi_{k,s}=\bP(s)\bP(s+1)\ldots\bP(k)$ as $\Phi_s\triangleq\lim_{k\rightarrow \infty}\Phi_{k,s}$ for notational simplicity. Then,
the entries $\Phi_{k,s}(i,j)$ converges to $1/N$ as $k$ goes to $\infty$ with a geometric rate.
The limit matrix $\Phi_s$ is doubly stochastic and correspond to a uniform steady distribution for all $s$, i.e.,
\beq
\Phi_s=\frac{1}{N}\pmb{1}\pmb{1}^T.
\eeq
\label{themorem_convergence_Phi}
\end{lem}

\begin{lem}[Lemma 4 in \cite{nedic2009distributed}]
Assume that $\bP(k)$ is doubly stochastic for all $k$.  Under Assumption~\ref{assumption-connectivity}, the difference between $1/N$ and any element of   $\Phi_{k,s}=\bP(s)\bP(s+1)\ldots\bP(k)$ can be bounded by
\beq
\left|\frac{1}{N}-\Phi_{k,s}(i,j)\right|\leq 2 \frac{(1+\beta^{-NB})}{1-\beta^{NB}}(1-\beta^{NB})^{(k-s)/NB},
\eeq
where  $\beta$ is the smallest positive value of all consensus matrices, i.e., $\beta=\arg\min P_{i,j}(k), \forall k$ with $P_{i,j}(k)>0, \forall i,j.$
\label{lemma_bound_Phi}
\end{lem}

\section{Additional Experimental Results}\label{app:sim}

In this section, we provide the details of the experiment setting in Section~\ref{sec:sim}. 

 \begin{wrapfigure}{rt}{0.35\linewidth}
	\centering
	\includegraphics[width=0.35\textwidth]{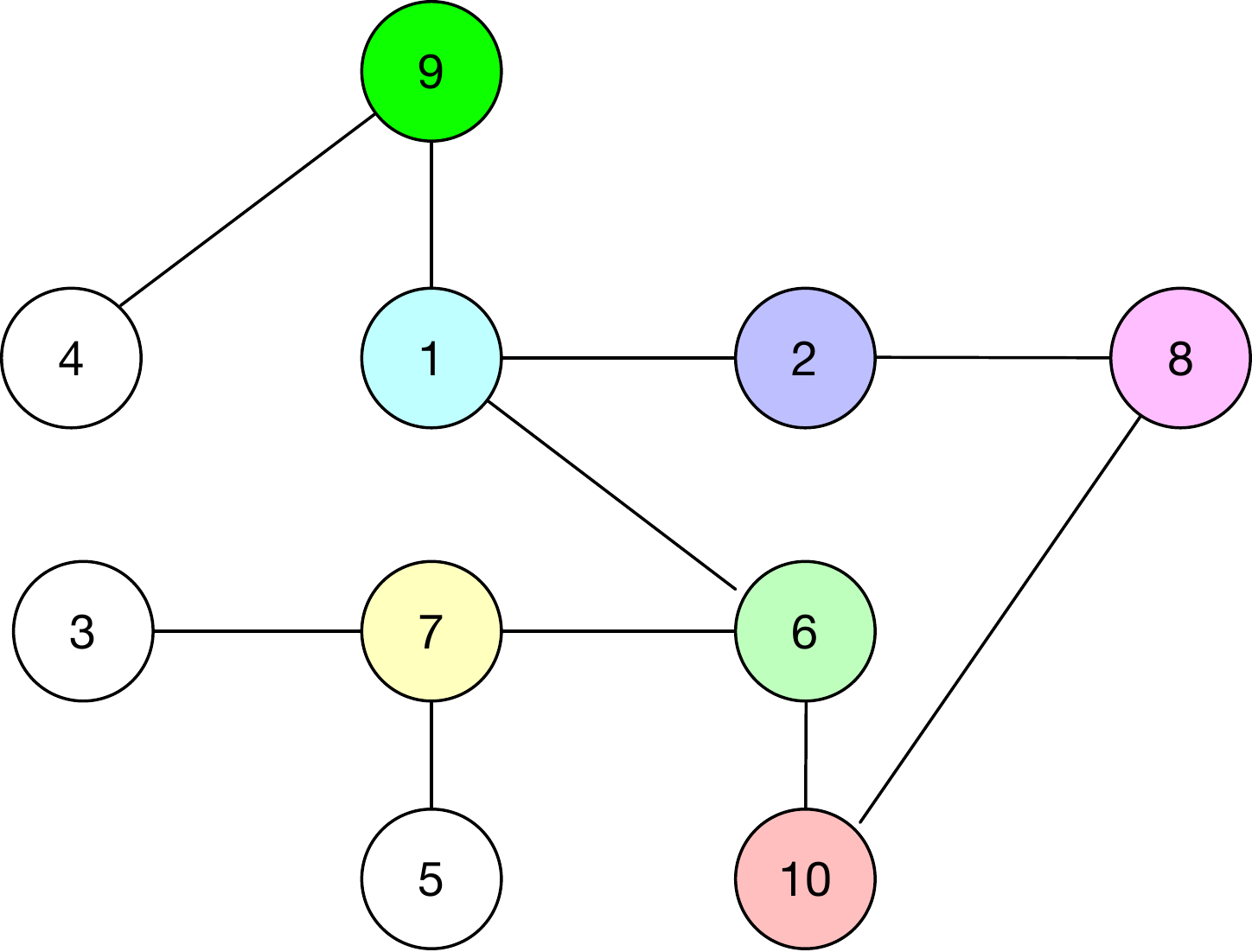}
	\caption{A connected network.}
	\label{fig:topology}
	\end{wrapfigure}

We evaluate \texttt{cb-DyBW} and \texttt{cb-Full} on the multi-class classification problem.  We use two models including the Logistic Regression Model (LRM) and a fully-connected neural network with 2 hidden layers (2NN) as shown in Table~\ref{tbl:2nn}, with MNIST \cite{lecun1998gradient,minst} and CIFAR-10 \cite{krizhevsky2009learning,cifar} datasets.  The MNIST dataset contains handwritten digits with $60,000$ samples for training and $10,000$ samples for testing.  The CIFAR-10 dataset consists of $60,000$ $32\times32$ color images in $10$ classes where $50,000$ samples are for training and the other $10,000$ samples for testing.  In addition to the results on a randomly connected graph with 6 workers presented in the main paper, we consider a randomly generated connected graph with 10 workers as shown in Figure~\ref{fig:topology}.  The loss functions we consider are the cross-entropy one for the LRM and the mean squared error (MSE) one for 2NN.  For both datasets, we evenly partition all training data among all workers, i.e., each worker observes $6,000$ data in MNIST and $5,000$ data in CIFAR-10.

\begin{table}[H]
	\centering
	\begin{tabular}{cc}
		\hline
		\text{Layer Type}  & \text{Size} \\ \hline
		\text{Fully Connected + ReLU}        & $256\times 256$ \\ 
		\text{Fully Connected + ReLU}        &$256\times 256$ \\
		\text{Fully Connected + SoftMax}   & $256\times 10$\\
		\hline
	\end{tabular}
	\caption{2NN architecture for MNIST.}
	\label{tbl:2nn}
\end{table}

\begin{figure}[h]
	\centering
	\includegraphics[width=0.95\textwidth]{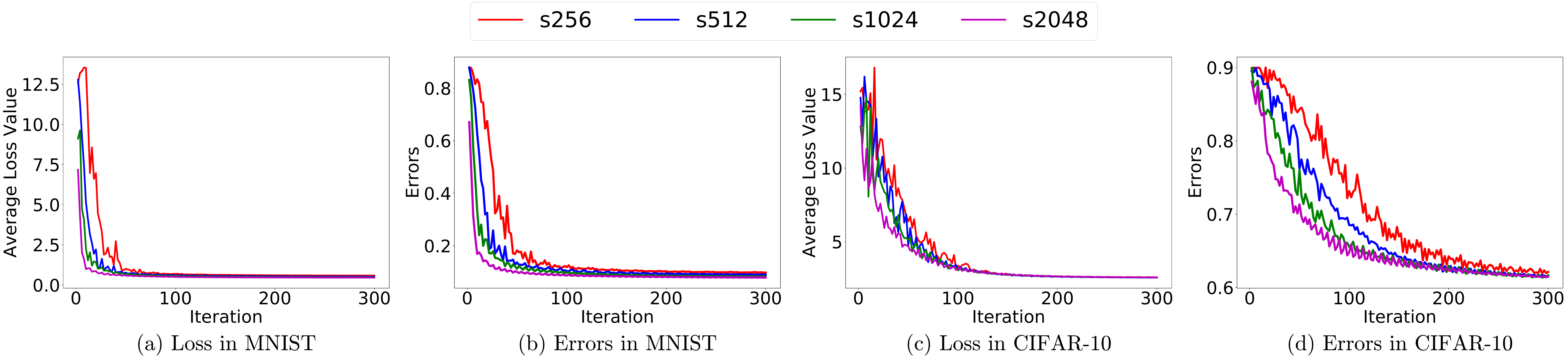}
	\caption{Impact of batch size.}
	\label{fig:batch}
\end{figure}

To enhance the training efficiency, we reduce the dimensions of MNIST (the dimension of samples is $784$) and CIFAR-10 (the dimension of samples is $3,072$) through the widely used principal component analysis (PCA) \cite{wold1987principal}. The learning rate is perhaps the most critical hyperparameter in distributed ML optimization problems.  A proper value of the learning rate is important; however, it is in general hard to find the optimal value.  The standard recommendation is to have the learning rate proportional to the aggregate batch size with full worker participation \cite{goyal2017accurate,smith2018don}.  With the consideration of dynamic backup workers, it is reasonable to adaptively set the learning rate \cite{chen2016revisiting,xu2020dynamic}. 
Hence, we choose $\eta(k)=\eta_0\cdot\delta^k$ where $\eta_0=1$ and $\delta=0.95.$  
Finally, batch size is another important hyperparameter, which is limited by the memory and computational resources available at each worker, or determined by generalization performance of the final model \cite{hoffer2017train}.   We test the impact of batch size using these two datasets and find that $1,024$ is a proper value (see Figure~\ref{fig:batch}).  This is because as we increase the batch size, the marginal improvement decreases.  From Figure~\ref{fig:batch}, we see that the performance with batch size $1,024$ is close to that of $2,048,$ however, each iteration takes shorter time with batch size $1,024$ than that with $2,048.$  Therefore, we use the batch size of $1,024$ in our experiments.   

\begin{figure}[h]
	\centering
	\includegraphics[width=0.95\textwidth]{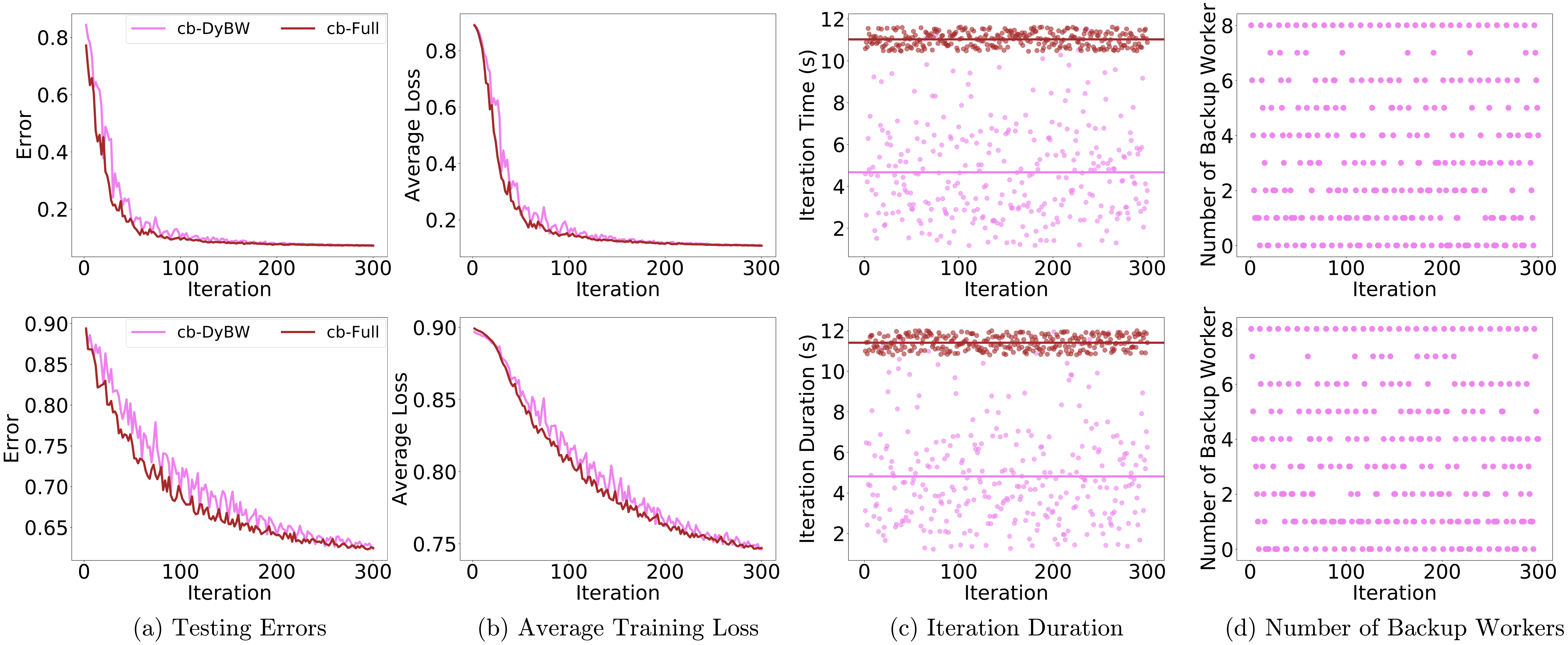}
	\caption{Performance of \texttt{cb-DyBW} and \texttt{cb-Full} for the 2NN model under MNIST (top) and CIFAR-10 (bottom). The straight line in (c) corresponds to the average number of backup workers over the iterations. }
	\label{fig:comparison-fnn}
\end{figure}

\begin{figure}[h]
	\centering
	\includegraphics[width=0.85\textwidth]{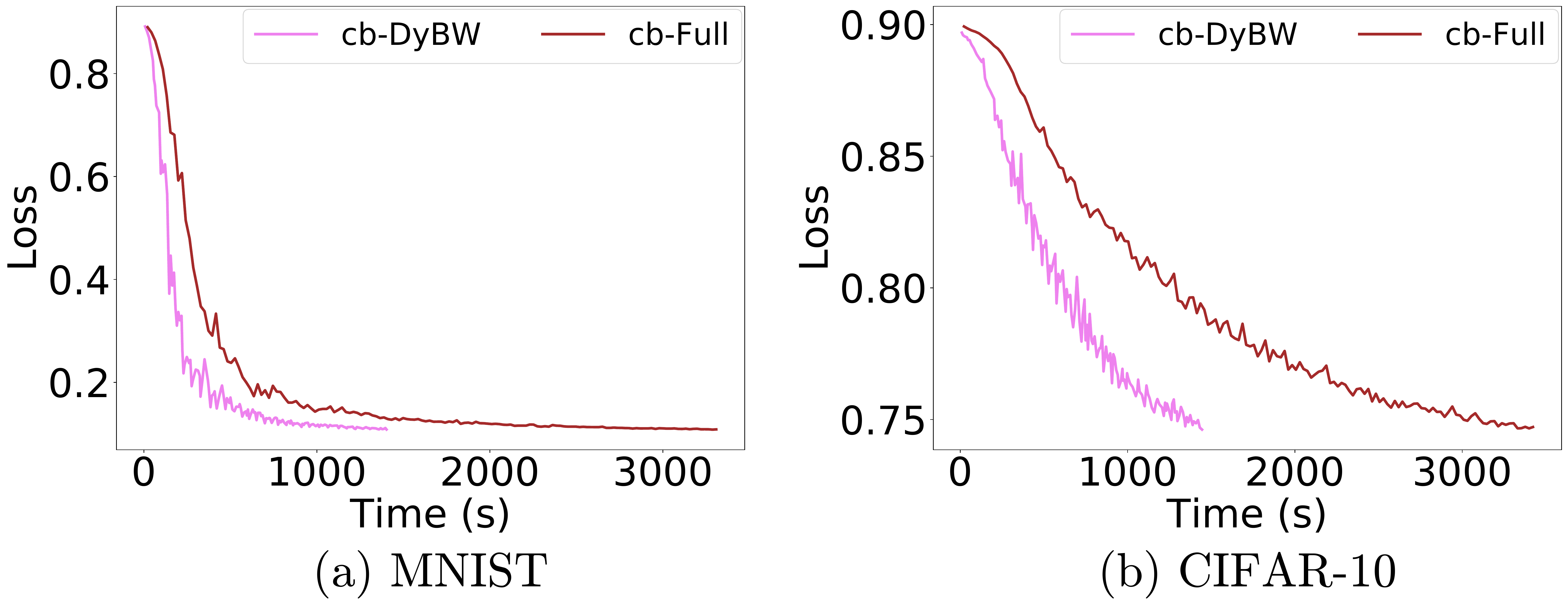}
	\caption{Loss versus time for the 2NN model under MNIST and CIFAR-10.}
	\label{fig:loss-time-fnn}
\end{figure}

Figure~\ref{fig:comparison-fnn} shows the (testing) errors, (training) loss and iteration duration of \texttt{cb-DyBW} and \texttt{cb-Full} for the 2NN model under MNIST (top) and CIFAR-10 (bottom) datasets, as well as the number of dynamic backup workers for \texttt{cb-DyBW}.  To better mimic the real-world situations with stragglers, we assume that there exists at least one straggler in each iteration in our experiments.  We observe that the number of iterations required for convergence is similar (in order sense) for both \texttt{cb-DyBW} and \texttt{cb-Full}, consistent with our theoretical results in Theorem \ref{thm:convergence}.  However, it is clear that \texttt{cb-DyBW} can dramatically reduce the duration of one iteration by 55\% on average compared to that of \texttt{cb-Full} from Figure~\ref{fig:comparison-fnn} (c).   This is because our proposed framework and algorithm \texttt{cb-DyBW} can dynamically and adaptively determine the number of backup workers for each worker during the training time so as to mitigate the effect of stragglers in the optimization problem.  As a result, \texttt{cb-DyBW} can significantly reduce the convergence time compared to \texttt{cb-Full} for a certain accuracy as observed in Figure~\ref{fig:loss-time-fnn}.  From Figure~\ref{fig:loss-time-fnn}, it is clear that our \texttt{cb-DyBW} only takes about 500 seconds to achieve a loss of 0.1 in MNIST while it takes more than 1,300 seconds for \texttt{cb-Full} to achieve a loss of 0.1, i.e., \texttt{cb-DyBW} dramatically reduces the convergence time by 62\%.  Similarly, \texttt{cb-DyBW} takes about 1,100 seconds to achieve a loss of 0.75, which takes more than 3,000 seconds for \texttt{cb-Full} to achieve the same accuracy, i.e.,  \texttt{cb-DyBW} dramatically reduces the convergence time by 63\%.
Finally, from Figure~\ref{fig:comparison-fnn} (d), it is clear that the number of backup workers is dynamically changing over time during the training time, which further validates our motivation and model.

\begin{figure}[h]
	\centering
	\includegraphics[width=0.95\textwidth]{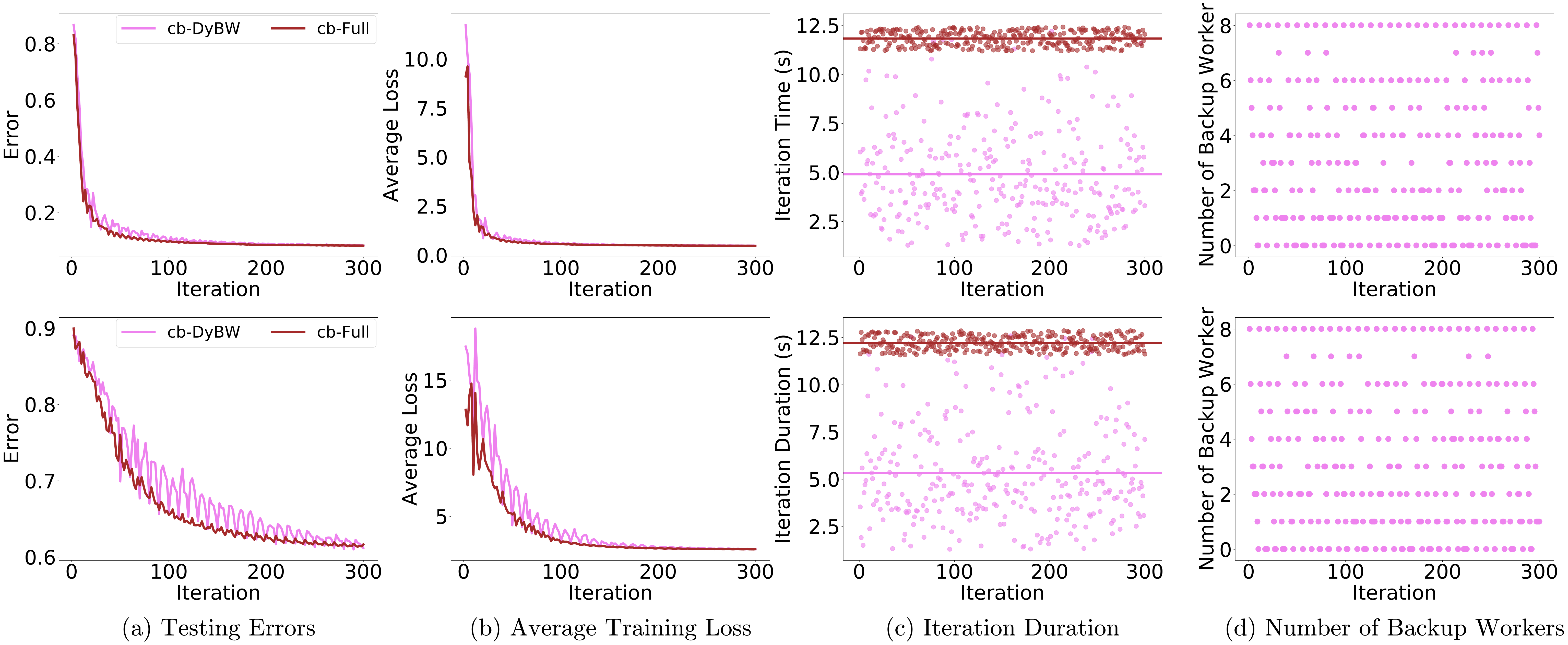}
	\caption{Performance of \texttt{cb-DyBW} and \texttt{cb-Full} for the LRM model under MNIST (top) and CIFAR-10 (bottom). The straight line in (c) corresponds to the average number of backup workers over the iterations. }
	\label{fig:comparison-lrm}
\end{figure}

\begin{figure}[h]
	\centering
	\includegraphics[width=0.85\textwidth]{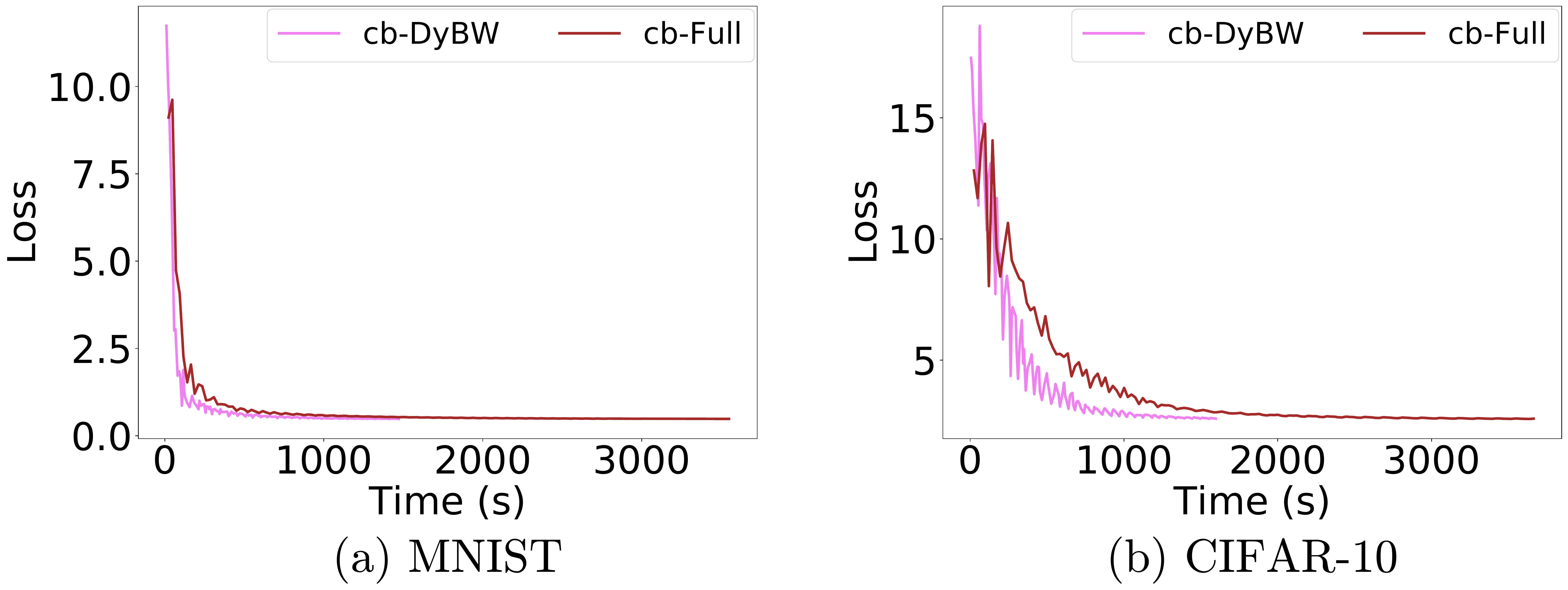}
	\vspace{-0.1in}
	\caption{Loss versus time for the LRM model under MNIST and CIFAR-10.}
	\label{fig:loss-time-lrm}
\end{figure}

Similarly, the performance of the (testing) errors, (training) loss and iteration duration of \texttt{cb-DyBW} and \texttt{cb-Full} for the LSM model under MNIST (top) and CIFAR-10 (bottom) datasets, the number of dynamic backup workers for \texttt{cb-DyBW}, as well as the convergence time are presented in Figures~\ref{fig:comparison-lrm} and~\ref{fig:loss-time-lrm}, respectively.  We can make similar observations.

\end{document}